\documentclass[10pt,twocolumn,letterpaper]{article}
\usepackage{cvpr}
\usepackage{times}
\usepackage{epsfig}
\usepackage{graphicx}
\usepackage{amsmath}
\usepackage{amssymb}
\usepackage{tabularx}
\usepackage{caption}
\usepackage[linesnumbered,ruled]{algorithm2e}

\cvprfinalcopy

\usepackage[pagebackref=true,breaklinks=true,letterpaper=true,colorlinks,bookmarks=false]{hyperref}

\def\cvprPaperID{4093} 

\DeclareMathOperator*{\argmin}{arg\,min}
\SetKwRepeat{Do}{do}{while}
\DontPrintSemicolon

\ifcvprfinal\fi
\begin{document}

\title{Deflecting Adversarial Attacks with Pixel Deflection}

\author{Aaditya Prakash, Nick Moran, Solomon Garber, Antonella DiLillo, James Storer\\
Brandeis University\\
{\tt\small \{aprakash,nemtiax,solomongarber,dilant,storer\}@brandeis.edu}
}

\maketitle

\begin{abstract}

CNNs are poised to become integral parts of many critical systems.
Despite their robustness to natural variations, image pixel values can be manipulated, via small, carefully crafted, imperceptible perturbations, to cause a model to misclassify images.
We present an algorithm to process an image so that classification accuracy is significantly preserved in the presence of such adversarial manipulations.
Image classifiers tend to be robust to natural noise, and adversarial attacks tend to be agnostic to object location.
These observations motivate our strategy, which leverages model robustness to defend against adversarial perturbations by forcing the image to match natural image statistics.
Our algorithm locally corrupts the image by redistributing pixel values via a process we term pixel deflection.
A subsequent wavelet-based denoising operation softens this corruption, as well as some of the adversarial changes. 
We demonstrate experimentally that the combination of these techniques enables the effective recovery of the true class, against a variety of robust attacks.  
Our results compare favorably with current state-of-the-art defenses, without requiring retraining or modifying the CNN. 
\end{abstract}

Code: \href{https://github.com/iamaaditya/pixel-deflection}{github.com/iamaaditya/pixel-deflection}
%






\section{Introduction}
\begin{figure}
   \includegraphics[width=1\linewidth]{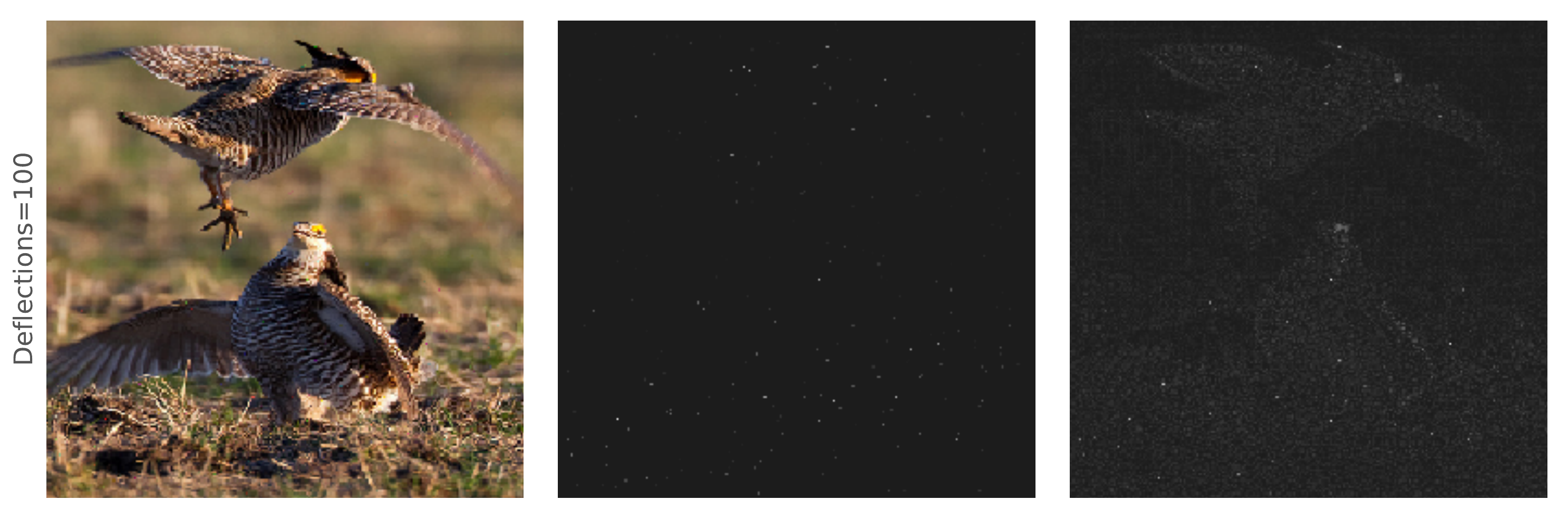}
   \includegraphics[width=1\linewidth]{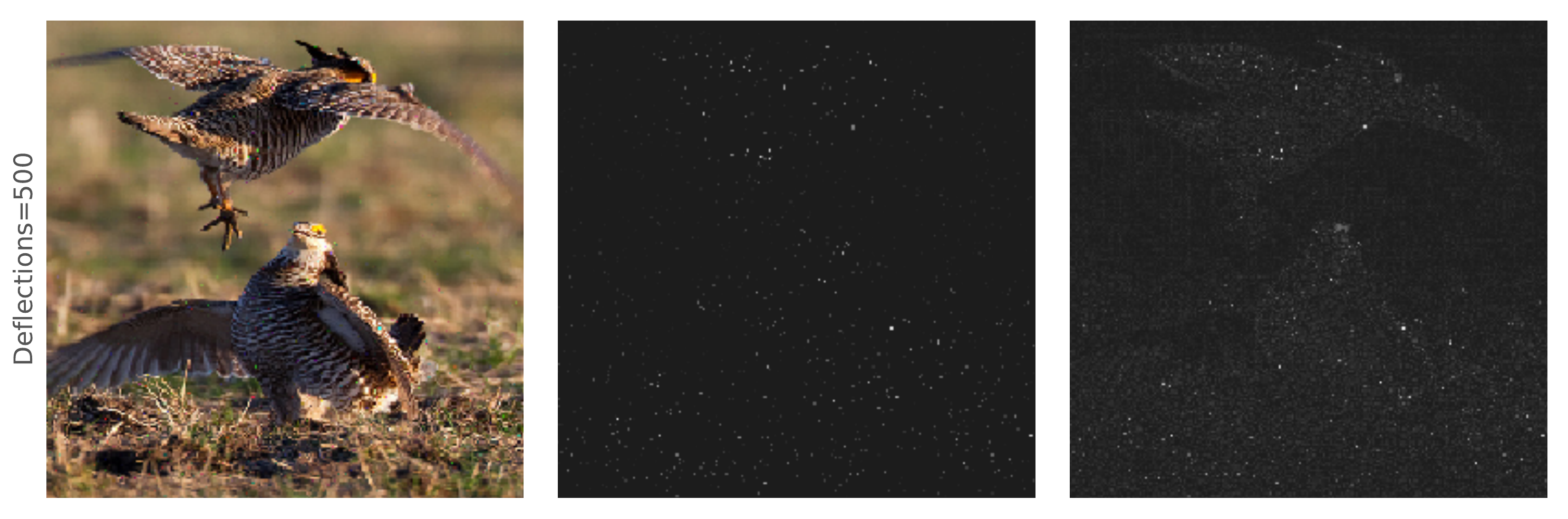}
   \includegraphics[width=1\linewidth]{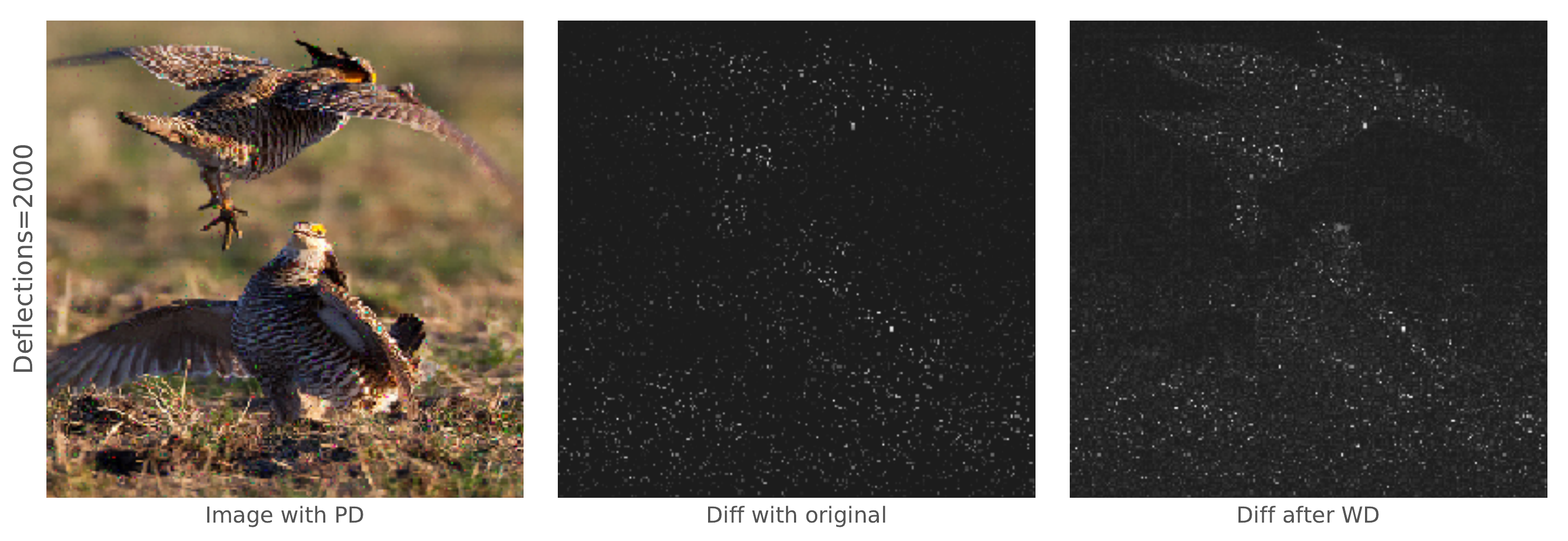}
   \label{fig:pixeldeflection}
   \caption{Impact of Pixel Deflection on a natural image and subsequent denoising using wavelet transform. Left: Image with given number of pixels deflected. Middle: Difference between clean image and deflected image. Right: Difference between clean image and deflected image after denoising. Enlarge to see details.}
   
\end{figure}

Image classification convolutional neural networks (CNNs) have become a part of many critical real-world systems.
For example, CNNs can be used by banks to read the dollar amount of a check~\cite{Bottou1997GlobalTO}, or by self-driving cars to identify stop signs~\cite{Papernot2016PracticalBA}.

The critical nature of these systems makes them targets for adversarial attacks.
Recent work has shown that classifiers can be tricked by small, carefully-crafted, imperceptible perturbations to a natural image.
These perturbations can cause a CNN to misclassify an image into a different class (\eg a ``1'' into a ``9'' or a stop sign into a yield sign). 

Thus, defending against these vulnerabilities will be critical to the further adoption of advanced computer vision systems.
Here, we consider \emph{white-box} attacks, in which an adversary can see the weights of the classification model.
Most of these attacks work by taking advantage of the differentiable nature of the classification model, \ie taking the gradient of the output class probabilities with respect to a particular pixel. 
Several previous works propose defense mechanisms that are differentiable transformations applied to an image before classification.
These differentiable defenses appear to work well at first, but attackers can easily circumvent these defenses by ``differentiating through them'', \ie by taking the gradient of a class probability with respect to an input pixel through both the CNN and the transformation. 

In this work, we present a defense method which combines two novel techniques for defending against adversarial attacks, which together modify input images in such a way that is (1) non-differentiable, and (2) frequently restores the original classification.
The first component, \emph{pixel deflection}, takes advantage of a CNN's resistance to the noise that occurs in natural images by randomly replacing some pixels with randomly selected pixels from a small neighborhood.
We show how to weight the initial random pixel selection using a \emph{robust activation map}.
The second approach, \emph{adaptive soft-thresholding} in the wavelet domain, which has been shown to effectively capture the distribution of natural images.
This thresholding process smooths adversarially-perturbed images in such a way so as to reduce the effects of the attacks.

Experimentally, we show that the combination of these approaches can effectively defend against state-of-the-art attacks~\cite{Szegedy2013IntriguingPO, Goodfellow2014ExplainingAH,Carlini2017TowardsET,MoosaviDezfooli2016DeepFoolAS,papernot2016limitations,Kurakin2016AdversarialEI}
Additionally, we show that these transformations do not significantly decrease the classifier's accuracy on non-adversarial images.

In Section~\ref{sec:advattacks}, we discuss the various attack techniques against which we will test our defense.  In Sections~\ref{sec:defenses} and~\ref{sec:related} we discuss the established defense techniques against which we will compare our technique.  In Sections~\ref{sec:pdrop},~\ref{sec:tpd} and~\ref{sec:wavelet} we lay out the components of our defense and provide the intuition behind them.  In Sections~\ref{sec:exp} and~\ref{sec:results}, we provide experimental results on a subset of ImageNet.

\section{Adversarial Attacks\label{sec:advattacks}}
It has been established that most image classification models can easily be fooled~\cite{Szegedy2013IntriguingPO,Goodfellow2014ExplainingAH}. 
Several techniques have been proposed which can generate an image that is perceptually indistinguishable from another image but is classified differently. 
This can be done robustly when model parameters are known, a paradigm called \textit{white-box attacks}~\cite{Goodfellow2014ExplainingAH,Kurakin2016AdversarialEI,Madry2017TowardsDL,Carlini2017TowardsET}.
In the scenario where access to the model is not available, called \textit{black-box attacks}, a secondary model can be trained using the model to be attacked as a guide.
It has been shown that the adversarial examples generated using these substitute models are transferable to the original classifiers~\cite{Papernot2016PracticalBA,Liu2016DelvingIT}.

Consider a given image $x$ and a classifier $F_\theta(\cdot)$ with parameters $\theta$.
Then an adversarial example for $F_\theta(\cdot)$ is an image $\hat{x}$ which is close
to $x$ (\ie $||x-\hat{x}||$ is small, where the norm used differs between attacks),  but the classifier's prediction for each of them is different, \ie $F(x) \neq F(\hat{x})$. 
\textit{Untargeted attacks} are  methods to produce such an image, given $x$ and $F_\theta(\cdot)$. 
\emph{Targeted attacks}, however, seek a $\hat{x}$ such that $F(\hat{x}) = \hat{y}$ for some specific choice of $\hat{y} \neq F(x)$, i.e. targeted attacks try to induce a specific class label, whereas untargeted attacks simply try to destroy the original class label.

Next, we present a brief overview of several well-known attacks, which form the basis for our experiments.

\paragraph{Fast Gradient Sign Method (FGSM)}\cite{Goodfellow2014ExplainingAH} is a single step attack process.
It uses the sign of the gradient of the loss function,  $\ell$,  \wrt to the image to find the adversarial perturbation. 
For a given value $\epsilon$, FGSM is defined as:
\begin{equation}
\hat{x} = x + \epsilon \text{sign} (\nabla \ell (F(x),x))
\end{equation}

\paragraph{Iterative Gradient Sign Method (IGSM)} \cite{Kurakin2016AdversarialEI} is an iterative version of FGSM. After each iteration the generated image is clipped to be within a $\epsilon L_\infty$ neighborhood of the original and this process stops when an adversarial image has been discovered. 
Both FGSM and IGSM minimize the $L_\infty$ norm \wrt to the original image. Let $x_0' = x$, then after $m$ iterations, the adversarial image is obtained by:
\begin{equation}
x_{m+1}' = \text{Clip}_{x,\epsilon} \Bigl\{x_m' + \alpha \times \text{sign}(\nabla \ell (F(x'_m),x'_m))  \Bigr\} 
\end{equation}

\paragraph{L-BFGS}~\cite{Szegedy2013IntriguingPO} tries to find the adversarial input as a box-constraint minimization problem.
L-BFGS optimization is used to minimize $L_2$ distance between the image and the adversarial example while keeping a constraint on the class label for the generated image.

\paragraph{Jacobian-based Saliency Map Attack (JSMA)} \cite{papernot2016limitations} estimates the saliency of each image pixel \wrt to the classification output, and modifies those pixels which are most salient. This is a targeted attack, and saliency is designed to find the pixel which increases the classifier's output for the target class while tending to decrease the output for other classes.

\paragraph{Deep Fool (DFool)}~\cite{MoosaviDezfooli2016DeepFoolAS} is an untargeted iterative attack. 
This method approximates the classifier as a linear decision boundary and then finds the smallest perturbation needed to cross that boundary.
This attack minimizes $L_2$ norm \wrt to the original image. 

\paragraph{Carlini \& Wagner (C\&W)}~\cite{Carlini2017TowardsET} is a recently proposed adversarial attack, and one of the strongest.
C\&W updates the loss function, such that it jointly minimizes $L_p$ and a custom differentiable loss function that uses the unnormalized outputs of the classifier (\textit{logits}). 
Let $Z_k$ denote the logits of a model for a given class $k$, and $\kappa$ a margin parameter. Then C\&W tries to minimize:
\begin{equation}
|| x - \hat{x} ||_p + c* max\left(Z(\hat{x}_y) - max\{Z(\hat{x})_k : k \neq y\},-\kappa\right)
\end{equation}
For our experiments, we use $L_2$ for the first term, as this makes the entire loss function differentiable and therefore easier to train. 
Limited success has been observed with $L_0$ and $L_\infty$ for images beyond CIFAR and MNIST.

We have not included recently proposed attacks like `Projected Gradient Descent'~\cite{Madry2017TowardsDL} and `One Pixel Attack'~\cite{Su2017OnePA} because although they have been shown to be robust on datasets of small images like CIFAR10 and MNIST, they do not scale well to large images. 
Our method is targeted towards large natural images where object localization is meaningful, i.e. that there are many  pixels outside the region of the image where the object is located.


\section{Defenses\label{sec:defenses}}

Given a classification model $F$ and an image $\tilde{x}$, which may either be an original image $x$, or an adversarial image $\hat{x}$, the goal of a defense method is to either augment either $F$ as $F'$ such that $F'(\tilde{x}) = F(x)$, or transform $\tilde{x}$ by a transformation $\mathcal{T}$ such that $F(\mathcal{T}(\tilde{x}))=F(x)$.

One method for augmenting $F$ is called Ensemble Adversarial training~\cite{Tramr2017EnsembleAT},  which augments the training of  deep convolutional networks to include various potential adversarial perturbations. 
This expands the decision boundaries around training examples to include some nearby adversarial examples, thereby making the task of finding an adversary within a certain $\epsilon$ harder than conventional models.
Another popular technique uses distillation from a larger network by learning to match the 
softmax~\cite{Papernot2016DistillationAA}.
This provides smoother decision boundaries and thus makes is harder to find an adversarial example which is imperceptible.
There are methods that proposes to detect the adversarial images as it passes through the classifier model~\cite{Meng2017MagNetAT,akhtar2017defense}.

Most transformation-based defense strategies suffer from accuracy loss with clean images ~\cite{Dziugaite2016ASO,Kurakin2016AdversarialEI}, \ie they produce $F(\mathcal{T}(x)) \neq F(x)$.
This is an undesirable side effect of the transformation process, and we propose a transformation which tries to minimize this loss while also recovering the classification of an adversarial image. 
Detailed discussion on various kinds of transformation based defenses is provided in section~\ref{sec:related}.

\section{Related Work \label{sec:related}}


Transformation-based defenses are a relatively recent and unexplored development in adversarial defense. 
The biggest obstacle facing most transformation-based defenses is that the transformation degrades the quality of non-adversarial images, leading to a loss of accuracy.
This has limited the success of transformations as a practical defense, as even those which are effective at removing adversarial transformations struggle to maintain the model's accuracy on clean images.
Our work is most similar to Guo~\etal's~\cite{CounteringAIGuo17} recently proposed transformation of image by quilting and Total Variance Minimization (TVM). 
Image quilting is performed by replacing patches of the input image with similar patches drawn from a bank of images.
They collect one million image patches from clean images and use a $k$-nearest neighbor algorithm to find the best match.
Image quilting in itself does not yield satisfactory results, so it is augmented with TVM.
In Total Variance Minimization, a substitute image is constructed by optimization such that total variance is minimized.
Total variation minimization has been widely used~\cite{Getreuer2012RudinOsherFatemiTV} as an image denoising technique. 
Our method uses semantic maps to obtain a better pixel to update and our update mechanism does not require any optimization and thus is significantly faster.

Another closely related work is from Luo \etal~\cite{FoveationbasedMALuo2015}. They propose a foveation-based mechanism.
Using ground-truth data about object coordinates, they crop the image around the object, and then scale it back to the original size.

Our model shares the hypothesis that not all regions of the image are equally important to a classifier.
Further, foveation-based methods can be fooled by finding an adversarial perturbation within the object bounding box.
Our model does not rely on a ground-truth bounding box, and the stochastic nature of our approach means that it is not restricted to only modifying a particular region of the input.

Yet another similar work is from Xie~\etal~\cite{MitigatingAnon208}, in which they pad the image and take multiple random crops and evaluate ensemble classification. 
This method utilizes the randomness property that our model also exploits. 
However, our model tries to spatially define the probability of a presence of a perturbation and subsequently uses wavelet-based transform to denoise the perturbations.

\section{Pixel Deflection \label{sec:pdrop}}

Much has been written about the lack of robustness of deep convolutional networks in the presence of adversarial inputs~\cite{EasilyFNguyen2015DeepNN,IntriguingSzegedy2013}.
However, most deep classifiers are robust to the presence of natural noise, such as sensor noise~\cite{DirtyPODiamond2017}. 
\begin{algorithm}
	\SetKwInOut{Input}{Input}
    \SetKwInOut{Output}{Output}
    \Input{Image $I$, neighborhood size $r$}
    \Output{Image $I'$ of the same dimensions as $I$}
    \For{$i\gets0$ \KwTo $K$}{
    	Let $p_i \sim \mathcal{U}(I)$\\
        Let $n_i \sim \mathcal{U}(R^r_p \cap I)$\\
        $I'[p_i] = I[n_i]$
    }
    \caption{Pixel deflection transform}
    \label{alg:swap}
\end{algorithm}
We introduce a form of artificial noise and show that most models are similarly robust to this noise.
We randomly sample a pixel from an image, and replace it with another randomly selected pixel from within a small square neighborhood.  We also experimented with other neighborhood types, including sampling from a Gaussian centered on the pixel, but these alternatives were less effective.


We term this process \textit{pixel deflection}, and give a formal definition in Algorithm~\ref{alg:swap}.
Let $R^r_{p}$ be a square neighborhood with apothem $r$ centered at a pixel $p$.
Let $\mathcal{U}(R)$ be the uniform distribution over all pixels within $R$.
Let $I_p$ indicate the value of pixel $p$ in image $I$.


\begin{figure}
   \includegraphics[width=1\linewidth]{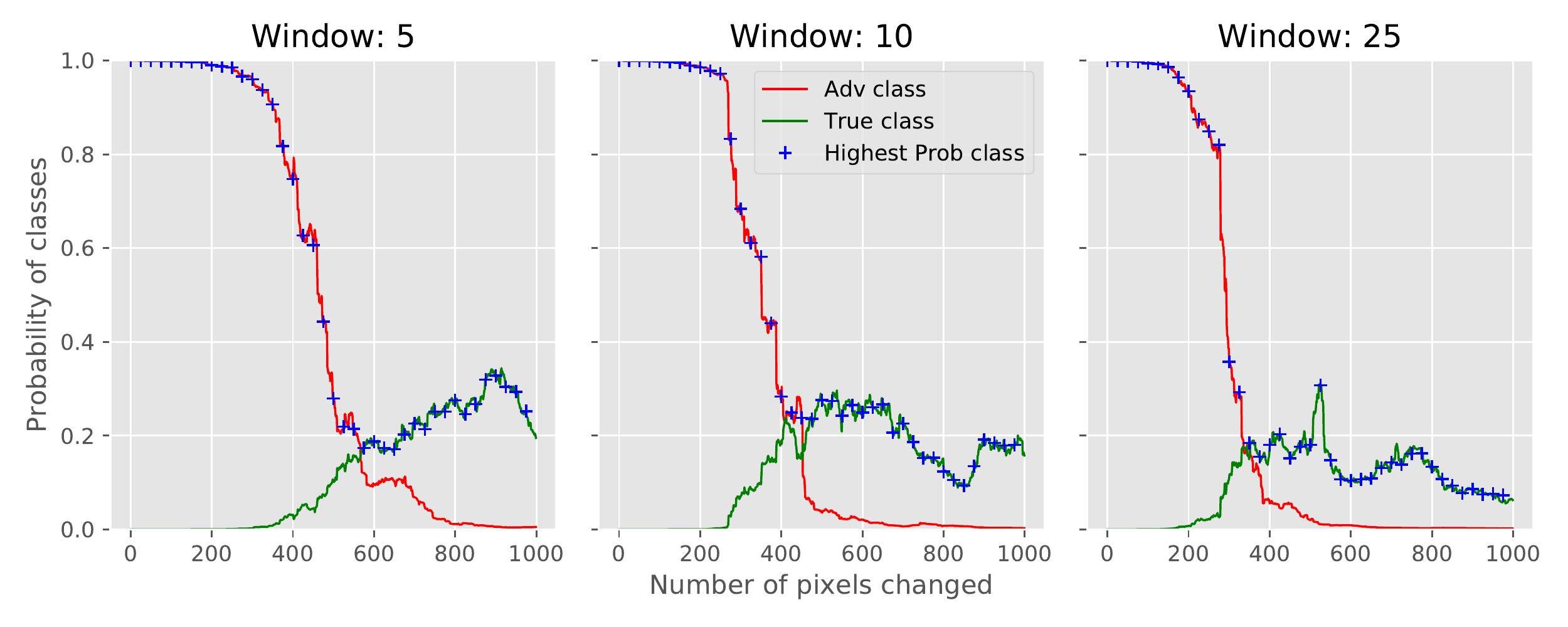}
   \includegraphics[width=1\linewidth]{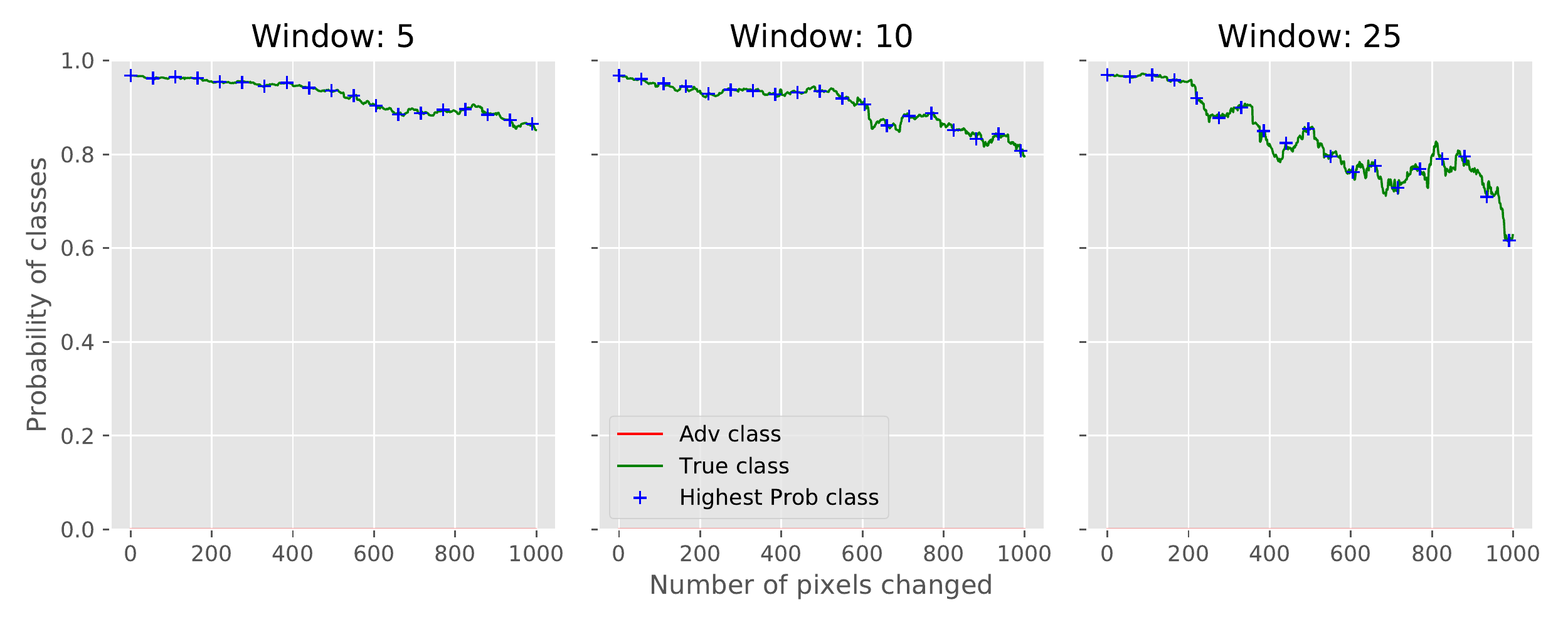}
   \caption{Average classification probabilities for an adversarial image (top) and clean image (bottom) after pixel deflection (Image size: 299x299)}
   \label{fig:pixelloss}
\end{figure}

As shown in Figure~\ref{fig:pixelloss}, even changing as much as $1\%$ (\ie $10$ times the amount changed in our experiments) of the original pixels does not alter the classification of a clean image.
However, application of pixel deflection enables the recovery of a significant portion of correct classifications.

\subsection{Distribution of Attacks}

Most attacks search the entire image plane for adversarial perturbations, without regard for the location of the image content.
This is in contrast with the classification models, which show high activation in regions where an object is present~\cite{Yosinski2015UnderstandingNN,Chattopadhyay2017GradCAMGG}. 
This is especially true for attacks which aim to minimize the $L_p$ norm of their changes for large values of $p$, as this gives little to no constraint on the total number of pixels perturbed.
In fact, Lou \etal \cite{FoveationbasedMALuo2015} use the object coordinates to mask out the background region and show that this defends against some of the known attacks. 
 

\begin{figure}
   \includegraphics[width=0.242\linewidth]{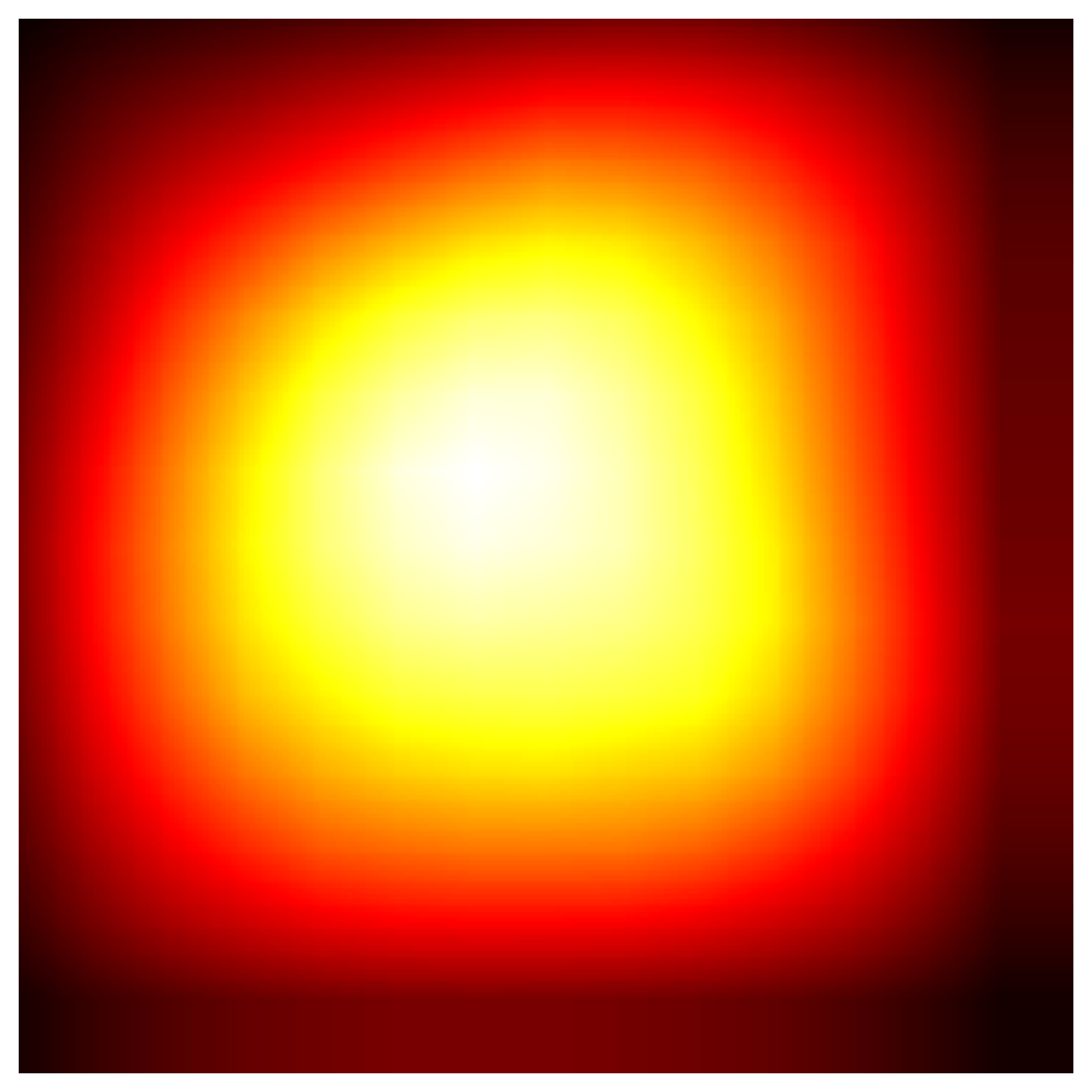}
   \includegraphics[width=0.242\linewidth]{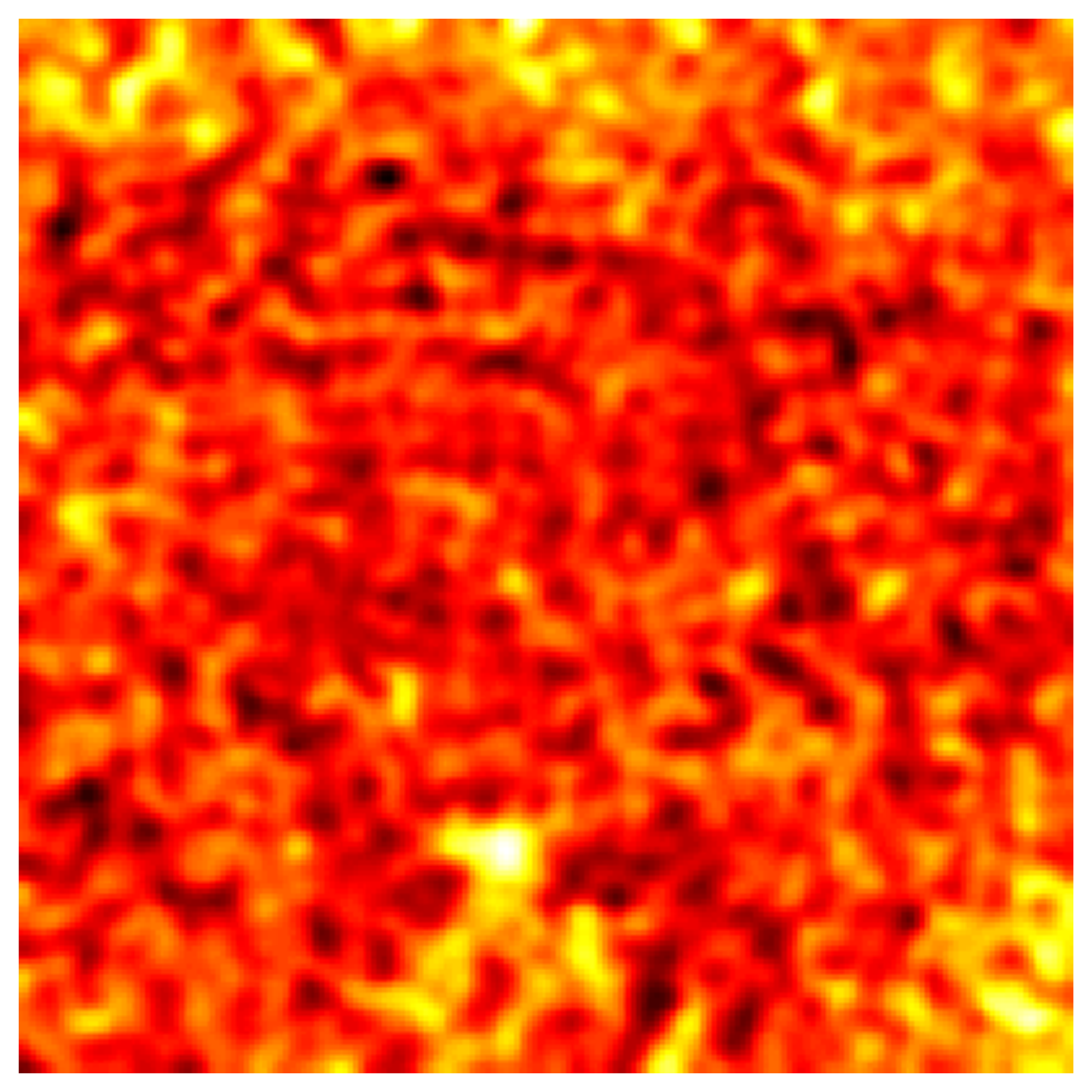}
   \includegraphics[width=0.242\linewidth]{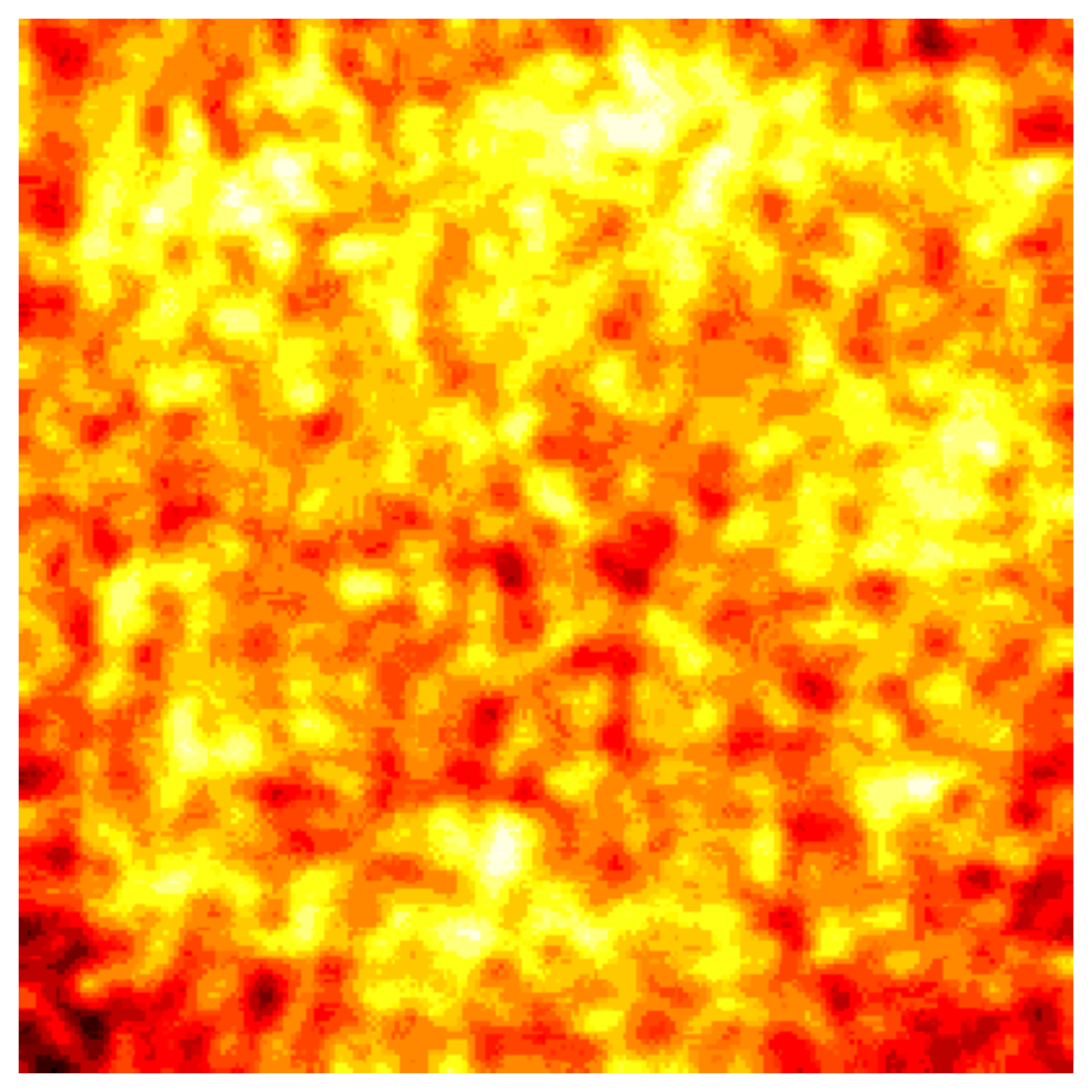}
   \includegraphics[width=0.242\linewidth]{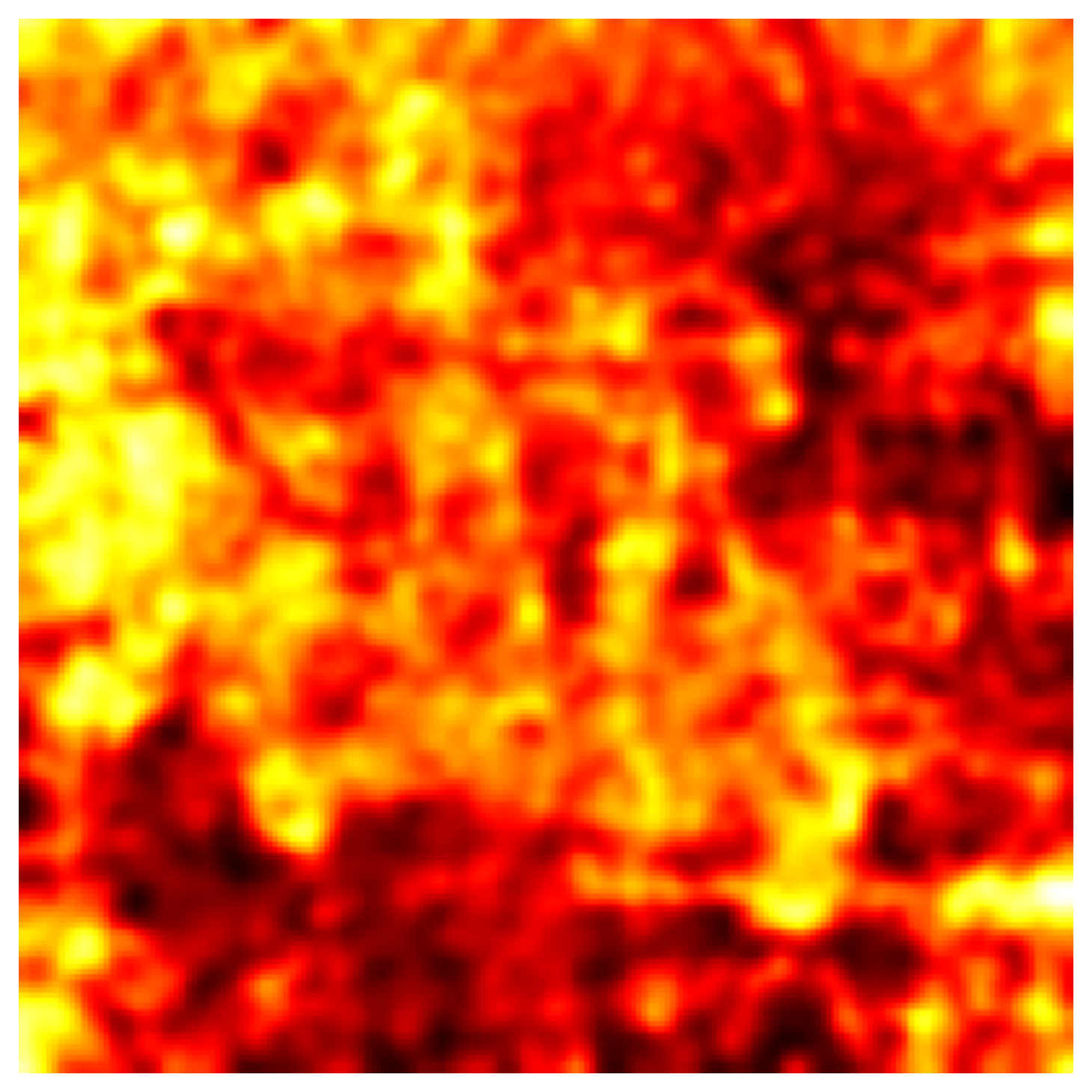}
   \includegraphics[width=0.242\linewidth]{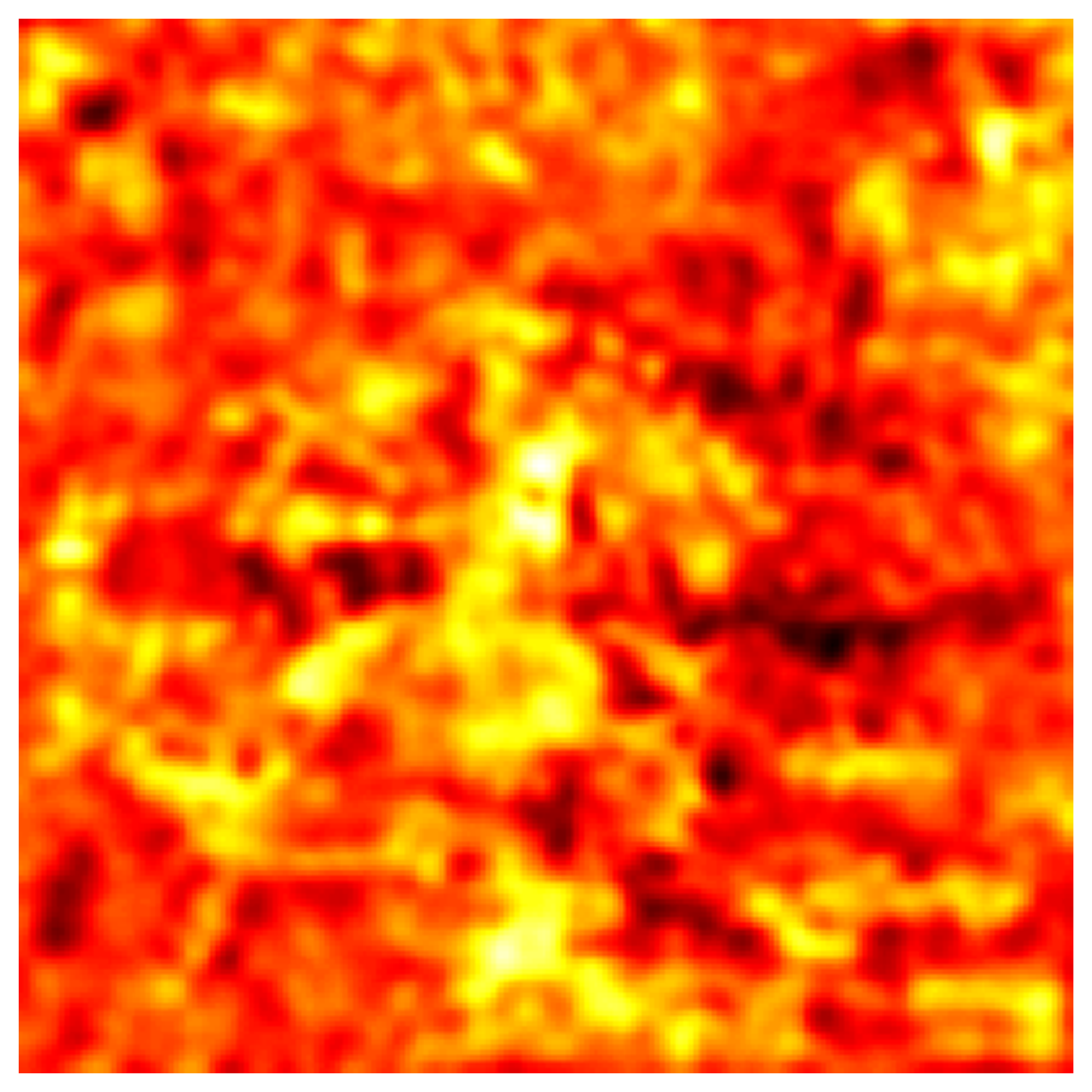}
   \includegraphics[width=0.242\linewidth]{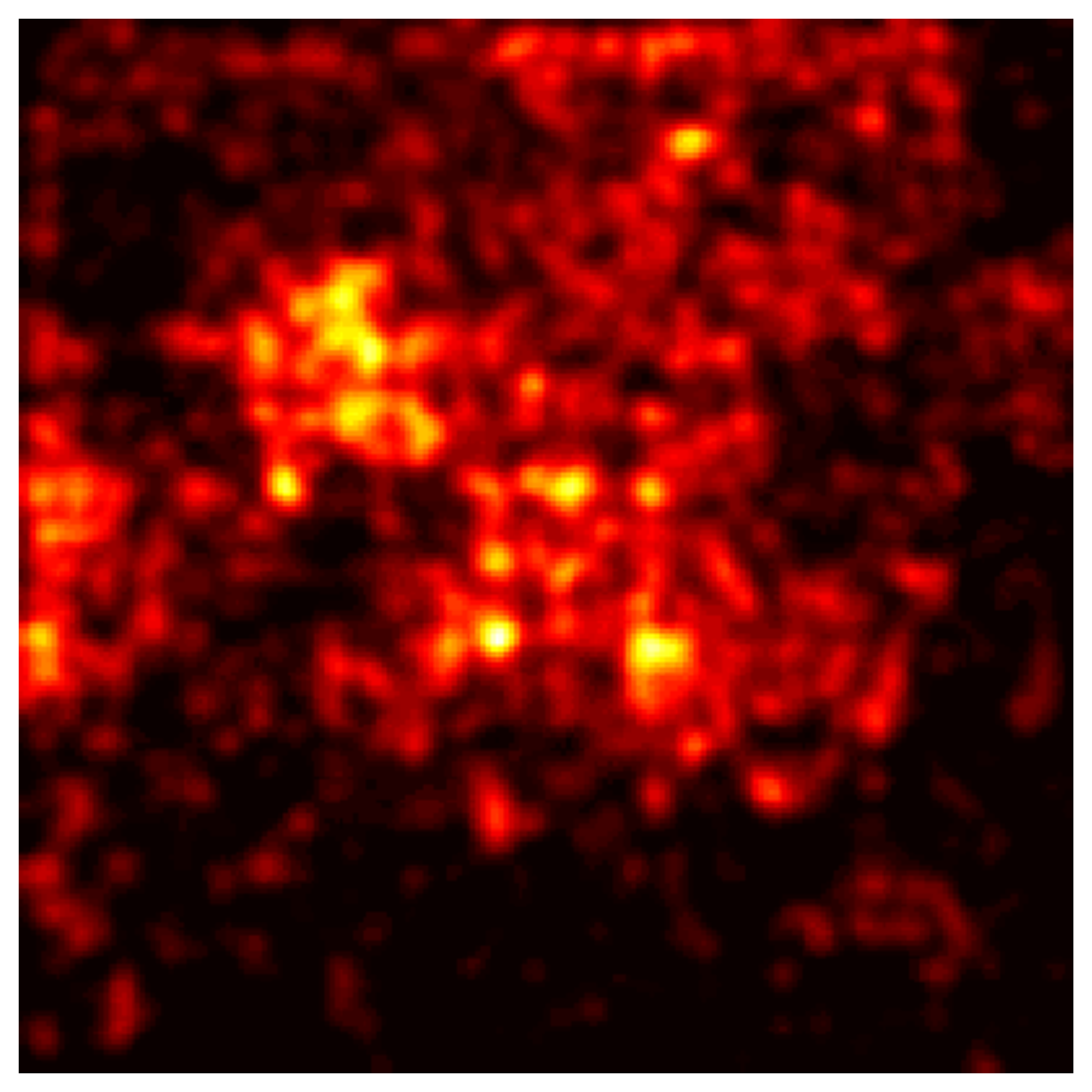}
   \includegraphics[width=0.242\linewidth]{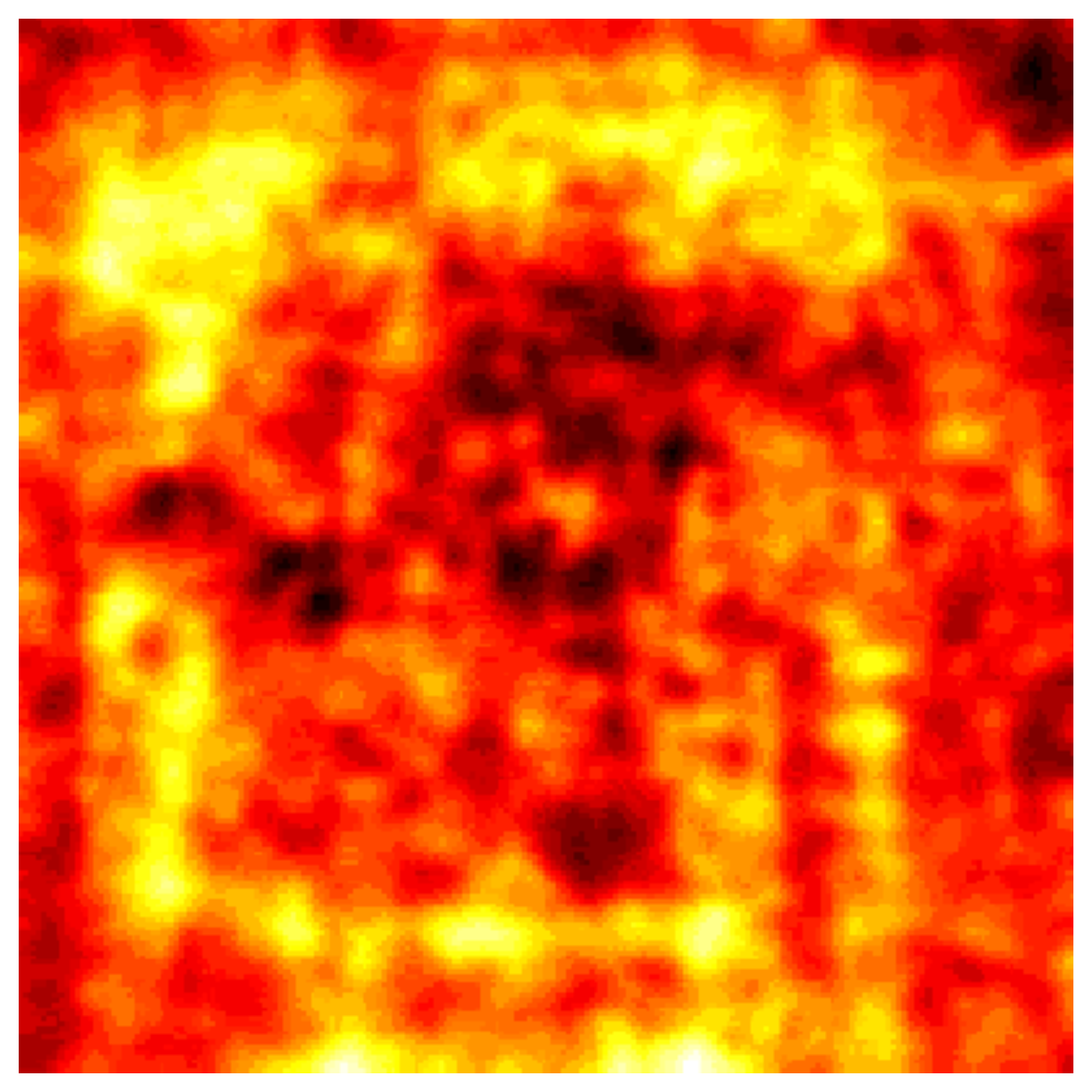}
   \includegraphics[width=0.242\linewidth]{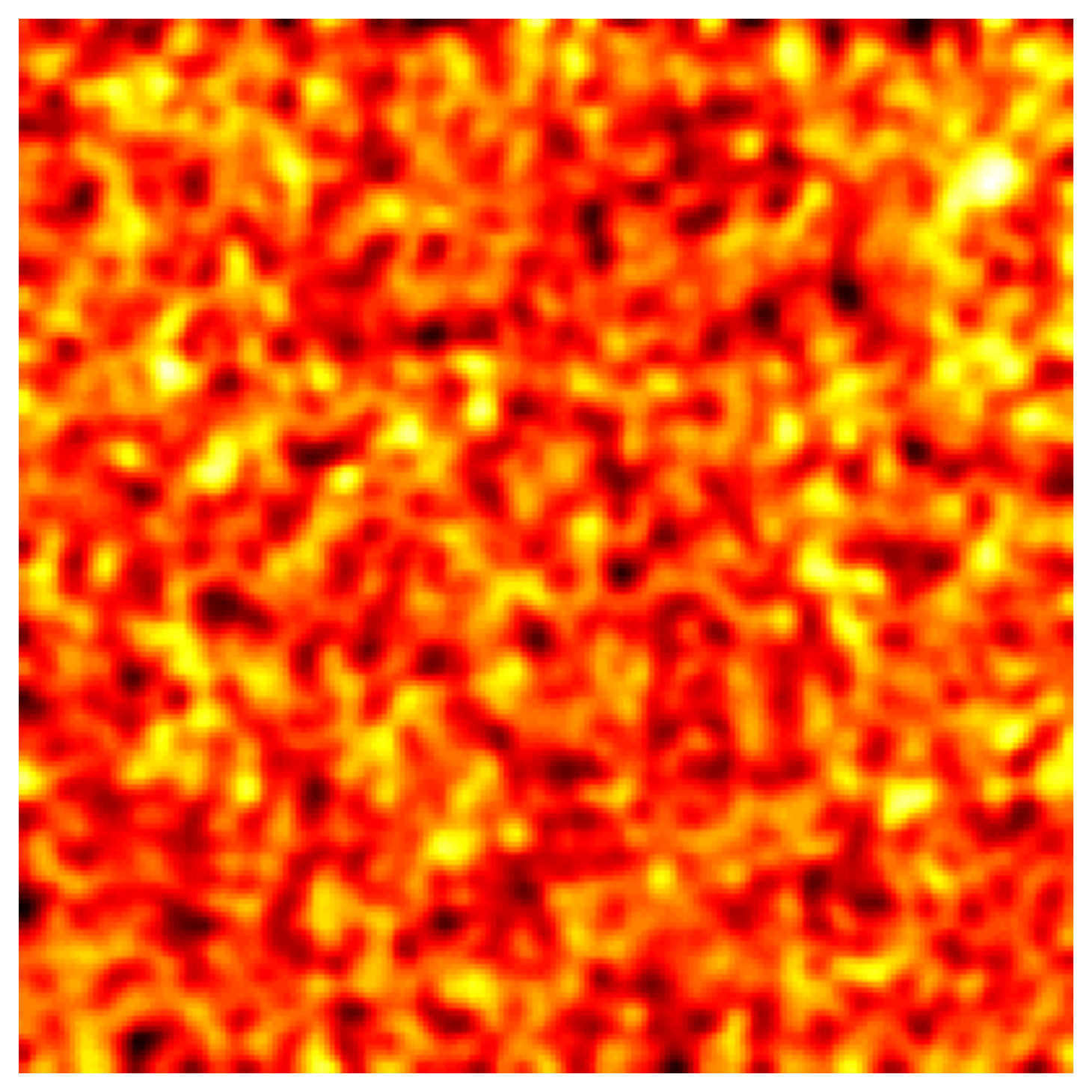}
   
   \caption{Visualization showing average location in the image where perturbation is added by an attacker.
   Clockwise from top left:
   Localization of most salient object in the image, FGSM, IGSM, FGSM-2 (higher $\epsilon$), Deep Fool, JSMA, LBFGS and Carlini-Wagner attack.
   \label{fig:visattack}}
   
\end{figure}

In Figure~\ref{fig:visattack} we show the average spatial distribution of perturbations for several attacks, as compared to the distribution of object locations (top left). 
Based on these ideas, we explore the possibility of updating the pixels in the image such that the probability of that pixel being updated is inversely proportional to the likelihood of that pixel containing an object.

\section{Targeted Pixel Deflection \label{sec:tpd}}

As we have shown in section~\ref{sec:pdrop}, image classification is robust against the loss of a certain number of pixels. 

In natural images, many pixels do not correspond to a relevant semantic object and are therefore not salient to classification.
Classifiers should then be more robust to pixel deflection if more pixels corresponding to the background are dropped as compared to the salient objects.
Luo \etal \cite{FoveationbasedMALuo2015} used this idea to mask the regions which did not contain the object,
however, their method has two limitations which we will seek to overcome.

First, it requires ground-truth object coordinates and it is, therefore, difficult to apply to unlabeled inputs at inference time. 
We solve this by using a variant of class activation maps to obtain an approximate localization for salient objects.
Class activation maps~\cite{CAMZhou2016LearningDF}  are a weakly-supervised localization~\cite{WeaklyOquab2015IsOL} technique in which
the last layer of a CNN, often a fully connected layer, is replaced with a global average pooling layer. 
This results in a heat map which lacks pixel-level precision but is able to approximately localize objects by their class.
We prefer to use weakly supervised localization over saliency maps~\cite{SALICONHuang2015}, as saliency maps are trained on human eye fixations 
and thus do not always capture object classes~\cite{DeepGazeKmmerer2014DeepGI}. 
Other weakly supervised localization techniques, such as regions-of-interest~\cite{Prakash2017SemanticPI}, capture more than a single object and thus are not suitable for improving single-class classification.

Second, completely masking out the background deteriorates classification of classes for which the model has come to rely on the co-occurrence of non-class objects.
For instance, airplanes are often accompanied by a sky-colored background, and most classifiers will have lower confidence when trying to classify an airplane outside of this context.
We take a Bayesian approach to this problem and use stochastic re-sampling of the background. 
This preserves enough of the background to protect classification and drops enough pixels to weaken the impact of adversarial input. 

\subsection{Robust Activation Map \label{sec:robustmap}}

Class activation maps~\cite{CAMZhou2016LearningDF} are a valuable tool for approximate semantic object localization.
Consider a convolutional network with $k$ output channels on the final convolution layer ($f$) with spatial dimensions of $x$ and $y$, and let $\boldsymbol{w}$ be a vector of size $k$ which is the result of applying a global max pool on each channel. 
This reduces channel to a single value, $\boldsymbol{w_k}$.
The class activation map, $M_c$ for a class $c$ is given by:
\begin{equation}
M_c(x,y) = \sum_k \boldsymbol{w}_k^c \; f_k(x,y)
\end{equation}
Generally, one is interested in the map for the class for which the model assigns the highest probability.
However, in the presence of adversarial perturbations to the input, the highest-probability class is likely to be incorrect.
Fortunately, our experiments show that an adversary which successfully changes the most likely class tends to leave the rest of the top-k classes unchanged.
Our experiments show that $38\%$ of the time the predicted class of adversarial images is the second highest class of the model for the clean image. 
Figure~\ref{fig:untarget} shows how the class of adversarial image relates to predictions on clean images.
ImageNet has one thousand classes, many of which are fine-grained. 
Frequently, the second most likely class is a synonym or close relative of the main class (\eg ``Indian Elephant'' and ``African Elephant'').
To obtain a map which is robust to fluctuations of the most likely class, we take an exponentially weighted average of the maps of the top-$k$ classes. 
\begin{equation}
	\widehat{M}(x,y) = \sum_i^k \frac{M_{c_i}(x,y)}{2^i}
\end{equation}
We normalize the map by diving it by its max so that values are in the range of $[0,1]$. 
Even if the top-1 class is incorrect, this averaging reduces the impact of mis-localization of the object in the image. 
\begin{figure}
   \includegraphics[width=1\linewidth]{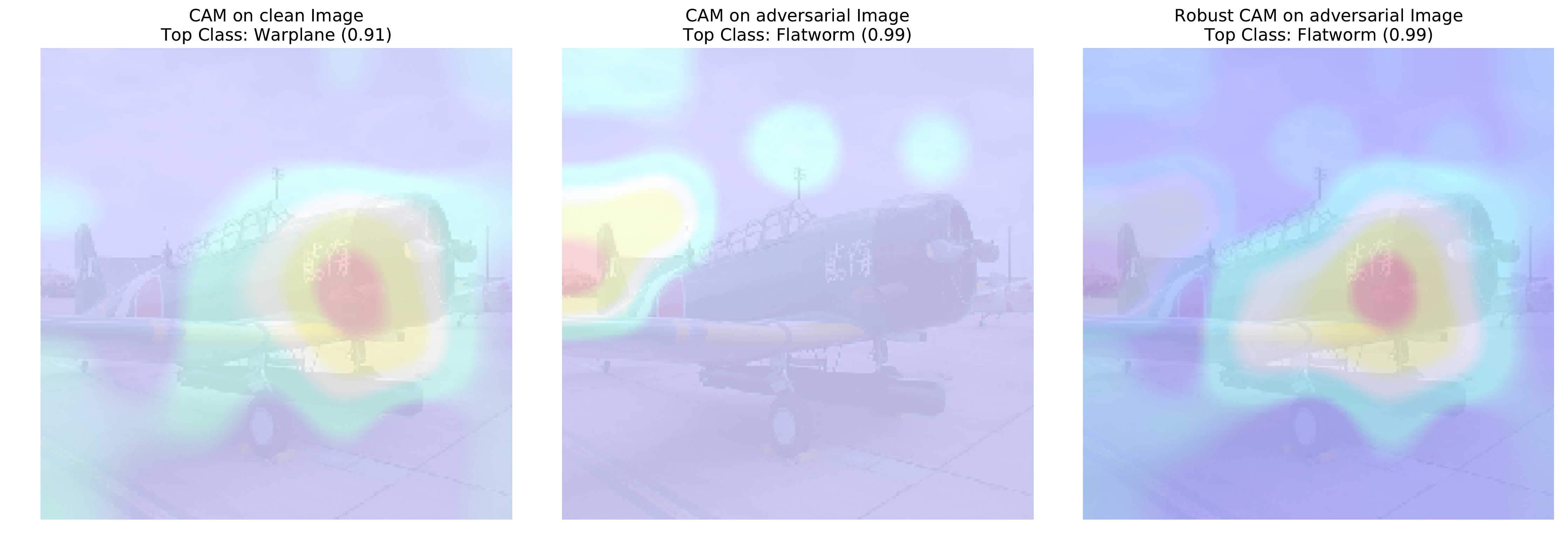}
   \caption{Difference between standard activation maps and robust maps under the presence of an adversary.}
   \label{fig:robustcam}
\end{figure}

The appropriate number of classes $k$ to average over depends on the total number of classes.
For ImageNet-1000, we used a fixed $k=5$.
While each possible class has its own class activation map (CAM), only a single robust activation map is generated for a particular image, combining information about all classes.
ImageNet covers wide variety of object classes and most structures found in other datasets are represented in ImageNet even if class names are not bijectional.
Therefore, Robust Activation Map (R-CAM) is trained once on ImageNet but can also localize objects from Pascal-VOC or Traffic Signs.

\section{Wavelet Denoising\label{sec:wavelet}}
Because both pixel deflection and adversarial attacks add noise to the image, it is desirable to apply a denoising transform to lessen these effects.
Since adversarial attacks do not take into account the frequency content of the perturbed image, they are likely to pull the input away from the class of likely natural images in a way which can be detected and corrected using a multi-resolution analysis. 

Works such as ~\cite{BayesShrinkChang2000,Simoncelli1999BayesianDO,field1987relations} have shown that natural images exhibit regularities in their wavelet responses which can be learned from data and used to denoise images.
These regularities can also be exploited to achieve better lossy image compression, the basis of JPEG2000.
Many vision and neuroscience researchers ~\cite{Marcelja1980MathematicalDO,Rust2005SpatiotemporalEO,Hubel1959ReceptiveFO} have suggested that the visual systems of many animals take advantage of these priors, as the simple cells in the primary visual cortex have been shown to have Gabor-like receptive fields.

Swapping pixels within a window will tend to add noise with unlikely frequency content to the image, particularly if the window is large.
This kind of noise can be removed by image compression techniques like JPEG, however, the quantization process in JPEG uses fixed tables that are agnostic to image content, and it quantizes responses at all amplitudes while the important image features generally correspond to large frequency responses.
This quantization reduces noise but also gets rid of some of the signal. 

Therefore, it is unsurprising that JPEG compression recovers correct classification on some of the adversarial images but  also reduces the classification accuracy on clean images \cite{Kurakin2016AdversarialEI,Das2017KeepingTB,Dziugaite2016ASO,CounteringAIGuo17}. 
Dziugaite \etal \cite{Dziugaite2016ASO} reported loss of $8$\% accuracy on clean images after undergoing JPEG compression. 

We, therefore, seek filters with frequency response better suited to joint space-frequency analysis than the DCT blocks (and more closely matching representations in the early ventral stream, so that features which have a small filter response are less perceptible) and quantization techniques more suited to denoising. 
Wavelet denoising uses wavelet coefficients obtained using \textit{Discrete Wavelet Transform} \cite{DWTantonini1992image}.
The wavelet transform represents the signal as a linear combination of orthonormal wavelets.
These wavelets form a basis for the space of images and are separated in space, orientation, and scale.
The Discrete Wavelet Transform is widely used in image compression~\cite{JPEG2000Adams2001} and image denoising~\cite{BayesShrinkChang2000,WaveletDenoisingRangarajan2002,Simoncelli1999BayesianDO}.

While the noise introduced by dropping a pixel is mostly high-frequency, the same cannot be said about the adversarial perturbations.
Several attempts have been made to quantify distribution of adversarial perturbations \cite{DropoutFeinman2017,ForesightLin2017DetectingAA} but recent work by Carlini and Wagner \cite{EasilyDetectedCarlini2017} has shown that most techniques fail to detect adversarial examples.
We have observed that for the perturbations added by well-known attacks, wavelet denoising yields superior results as compared to block DCT. 

\subsection{Hard \& Soft Thresholding}


The process of performing a wavelet transform and its inverse is lossless and thus does not provide any noise reduction.
In order to reduce adversarial noise, we need to apply thresholding to the wavelet coefficients before inverting the transform.
Most compression techniques use a hard thresholding process, in which all coefficients with magnitude below the threshold are set to zero: $Q(\hat{X}) = \hat{X} \; \forall \; |\hat{X}| > T_h$, where $\hat{X}$ is the wavelet transform of $X$, and $T_h$ is the threshold value.
The alternative is soft thresholding, in which we additionally subtract the threshold from the magnitude of coefficients above the threshold: $Q(\hat{X}) = \text{sign}(\hat{X})\times \text{max}(0,|\hat{X}| - T_h)$.
Jansen \etal \cite{ThresholdingJansen2012noise} observed that hard thresholding results in over-blurring of the input image, while soft thresholding maintains better PSNR.
By reducing all coefficients, rather than just those below the threshold, soft thresholding avoids introducing extraneous noise.
This allows our method to preserve classification accuracy on non-adversarial images.

\subsection{Adaptive Thresholding \label{sec:contentthresholding}}
Determining the proper threshold is very important, and the efficacy of our method relies on the ability to pick a threshold in an adaptive, image specific manner.
The standard technique for determining the threshold for wavelet denoising is to use a universal threshold formula called \textit{VisuShrink}. 
For an image $X$ with $N$ pixels, this is given by $\sigma\sqrt{2\log N}$, where $\sigma$ is the variance of the noise to be removed and is a hyper-parameter. 
However, we used \textit{BayesShrink} \cite{BayesShrinkChang2000}, which models the threshold for each wavelet coefficient as a Generalized Gaussian Distribution (GGD). 
The optimal threshold is then assumed to be the value which minimizes the expected mean square error \ie
\begin{equation}
T_h*(\sigma_x, \beta) = \argmin_{T_h} E(\hat{X} - X)^2 \; \approx \frac{\sigma^2}{\sigma_x}
\label{eqn:bayesshrink}
\end{equation}
where $\sigma_x$ and $\beta$ are parameters of the GGD for each wavelet sub-band.
In practice, an approximation, as shown on right side of equation~\ref{eqn:bayesshrink}, is used. 
This ratio, also called $T_{\text{Bayes}}$, adapts to the amount of noise in the given image.
Within a certain range of $\beta$ values, BayesShrink has been shown to effectively remove artificial noise while preserving the perceptual features of natural images~\cite{BayesShrinkChang2000,WaveletDenoisingRangarajan2002}.
As our experiments are carried out with images from ImageNet, which is a collection of natural images, we believe this is an appropriate thresholding technique to use.
Yet another popular thresholding technique is Stein's Unbiased Risk Estimator (SUREShrink), which computes unbiased estimate of $E(\hat{X} - X)^2 $. SUREShrink requires optimization to learn $T_h$ for a given coefficient. We empirically evaluated results and SUREShrink did not perform as well as BayesShrink. Comparative results are  shown in Table~\ref{tblshrink}.

\section{Method}

The first step of our method is to corrupt the adversarial noise by applying targeted pixel deflection as follows: 

(a) Generate a robust activation map $\widehat{M}$, as described in section~\ref{sec:robustmap}.

(b) Uniformly sample a pixel location $(x,y)$ from the image, and obtain the normalized activation map value for that location, $v_{x,y} = \widehat{M}(x,y)$.

(c) Sample a random value from a uniform distribution $\mathcal{U}(0,1)$. If $v_{x,y}$ is lower than the random value, we deflect the pixel using the algorithm shown in Algorithm ~\ref{alg:swap}. 

(d) Iterate this process $K$ times. \\

The following steps are used to soften the impact of pixel deflection:

(a) Convert the image to $YC_bC_r$ space to decorrelate the channels. $YC_bC_r$ space is perceptually meaningful and thus has similar denoising advantages to the wavelets.  

(b) Project the image into the wavelet domain using the discrete wavelet transform. We use the \textit{db1} wavelet, but similar results were obtained with \textit{db2} and \textit{haar} wavelets. 

(c) Soft threshold the wavelets using BayesShrink. 

(d) Compute the inverse wavelet transform on the shrunken wavelet coefficients.

(e) Convert the image back to RGB.

\section{Experimental Design\label{sec:exp}}

We tested our method on $1000$ randomly selected images from the ImageNet~\cite{Deng2009ImageNetAL} Validation set. 
We use ResNet-50~\cite{He2016DeepRL} as our classifier. 
We obtain the pre-trained weights from TensorFlow's GitHub repository.
These models achieved a Top-1 accuracy of $76\%$ on our selected images. 
This is in agreement with the accuracy numbers reported in~\cite{He2016DeepRL} for a single-model single-crop inference.

By the definition set by adversarial attacks, an attack is considered successful by default if the original image is already mis-classified. 
In this case, the adversary simply returns the original image unmodified. 
However, these cases are not useful for measuring the effectiveness of an attack or a defense as there is no pixel level difference between the images.
As such, we restrict our experiments to those images which are correctly classified in the absence of adversarial noise.
Our attack models are based on the Cleverhans~\cite{papernot2017cleverhans} library\footnote{https://github.com/tensorflow/cleverhans} with model parameters that aim to achieve the highest possible misclassification score with a normalized RMSE ($|\boldmath{L}_2|$) budget of $0.02-0.04$.

We will publicly release  our implementation code.

\subsection{Training}

\begin{figure}
   \includegraphics[width=0.49\linewidth]{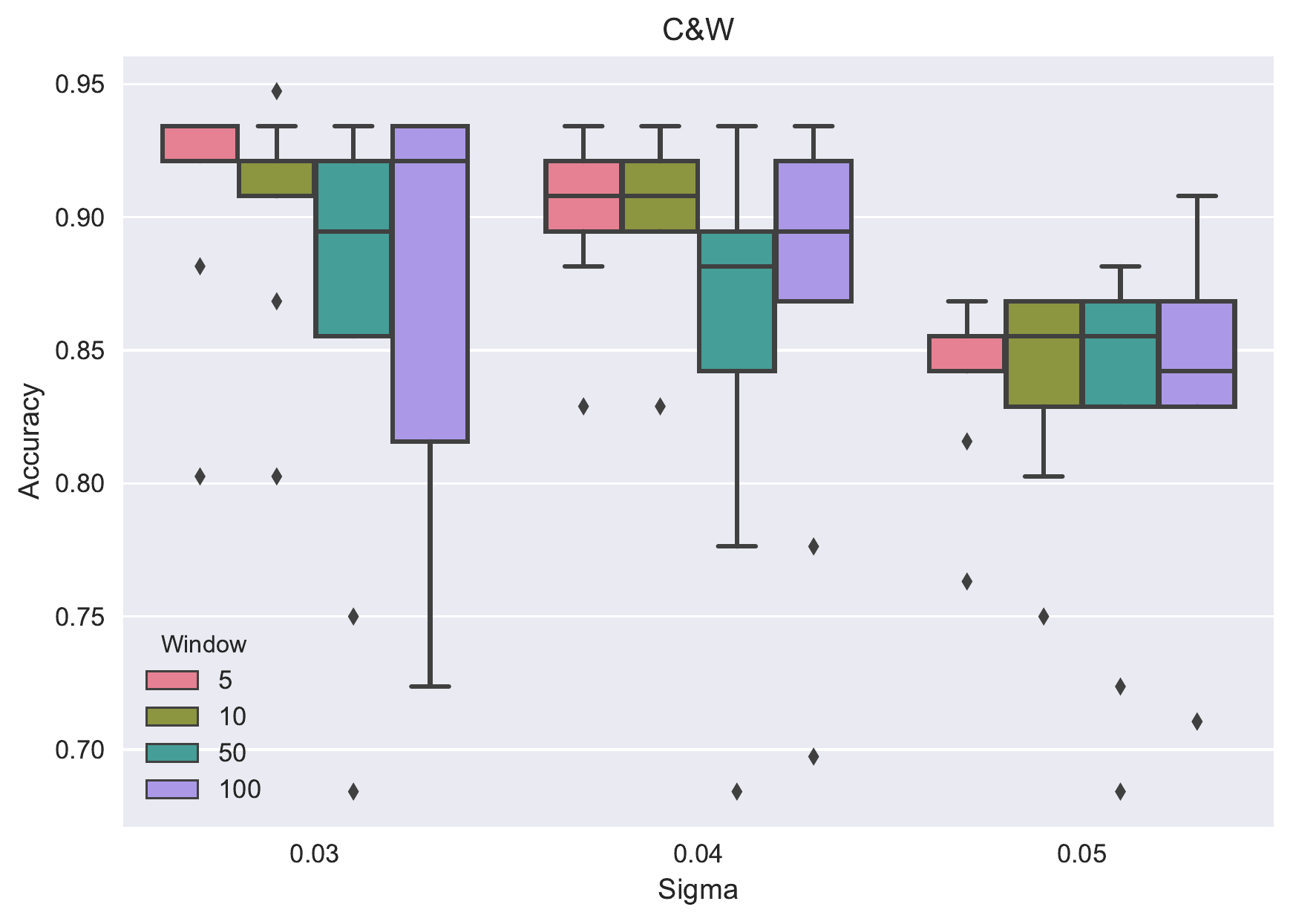}
   \includegraphics[width=0.49\linewidth]{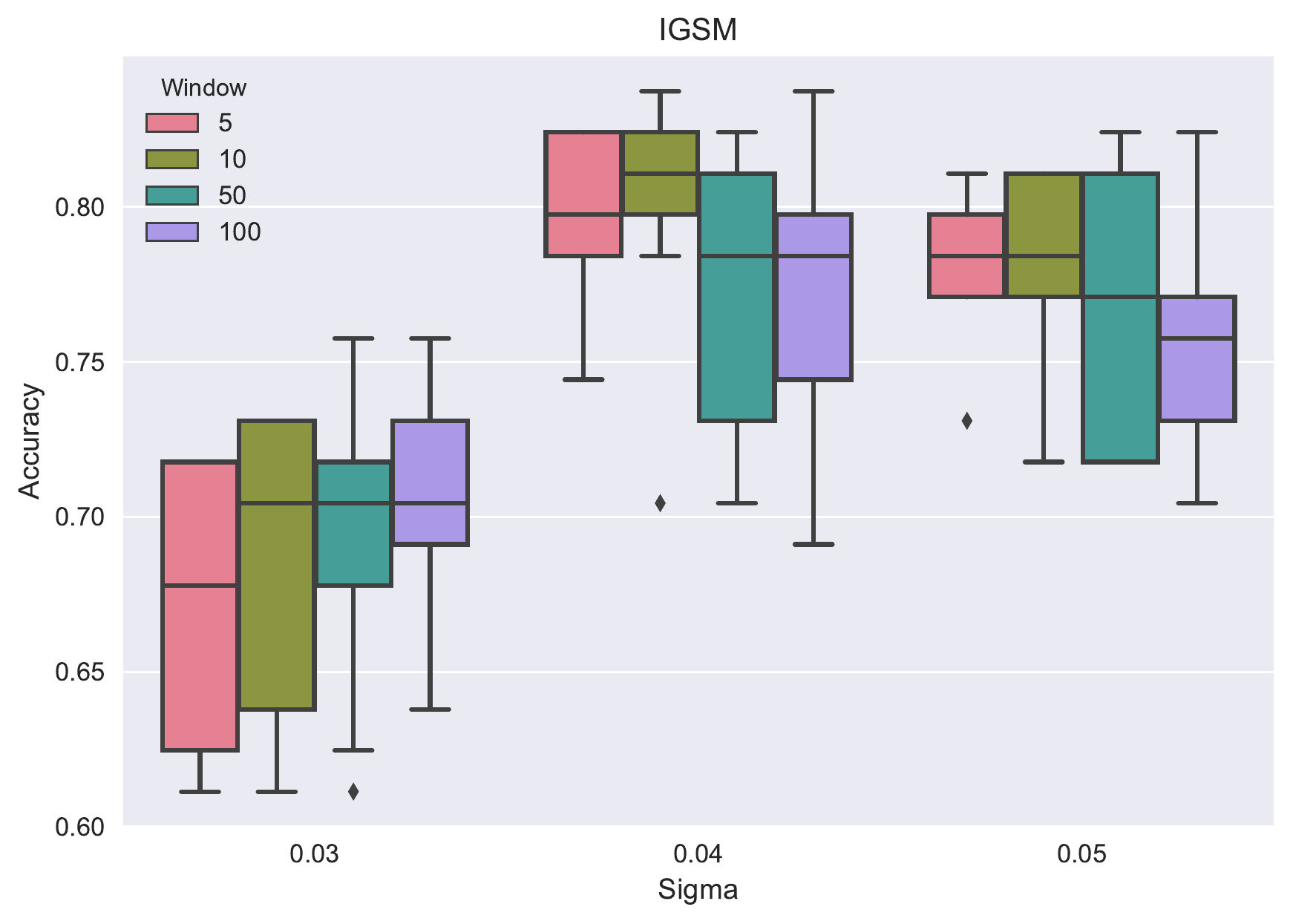}
   \includegraphics[width=0.49\linewidth]{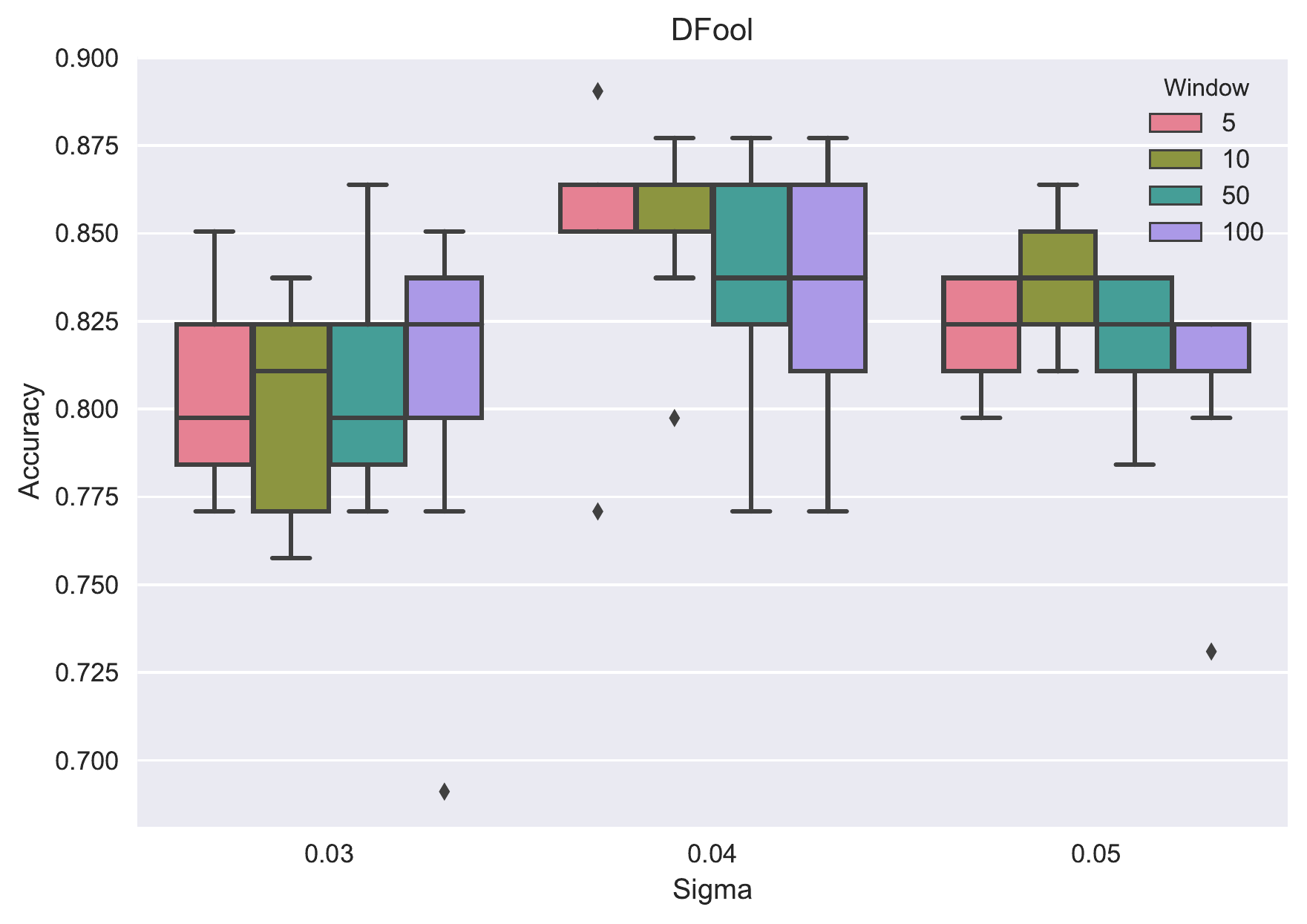}
   \includegraphics[width=0.49\linewidth]{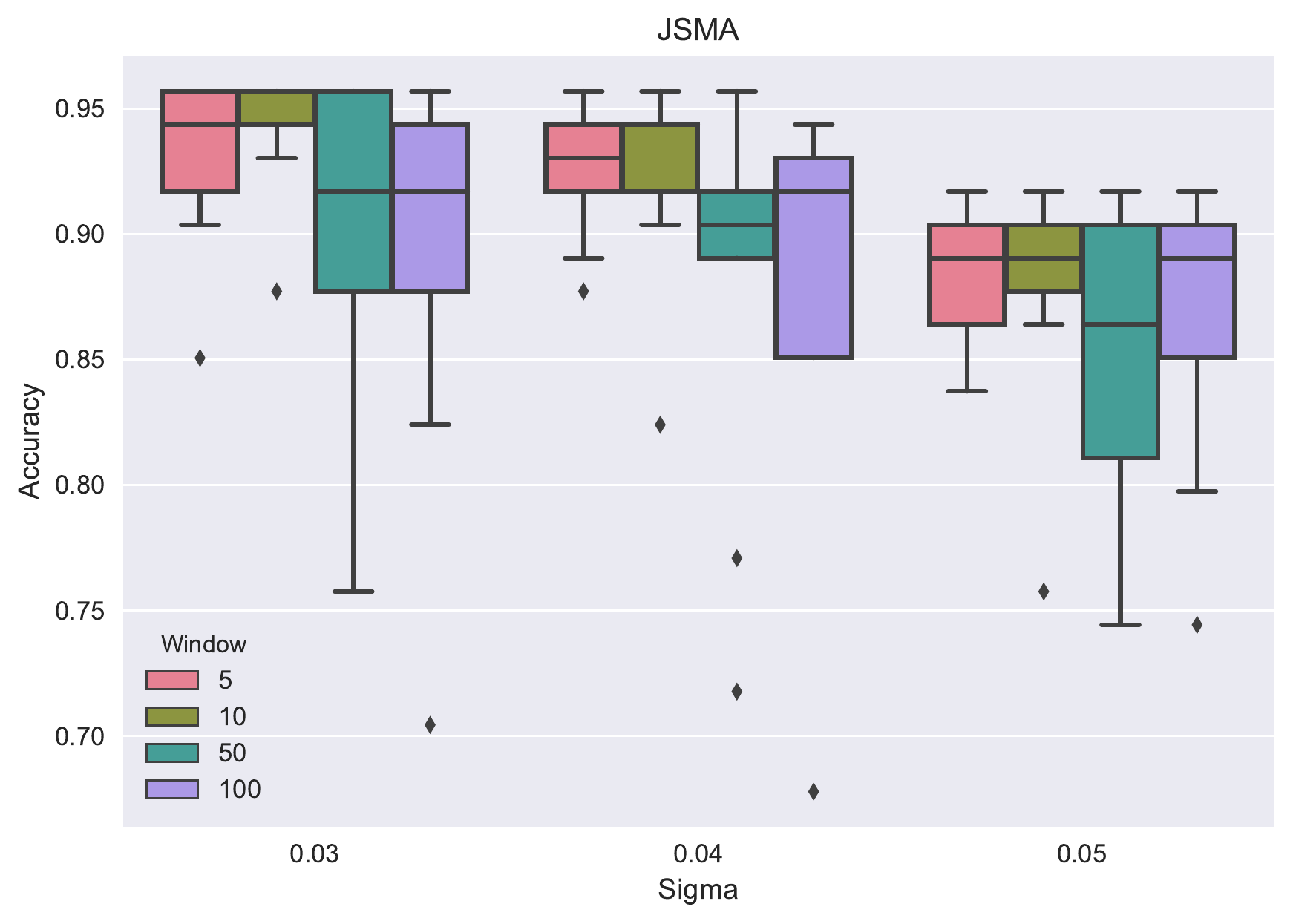}
   \includegraphics[width=0.49\linewidth]{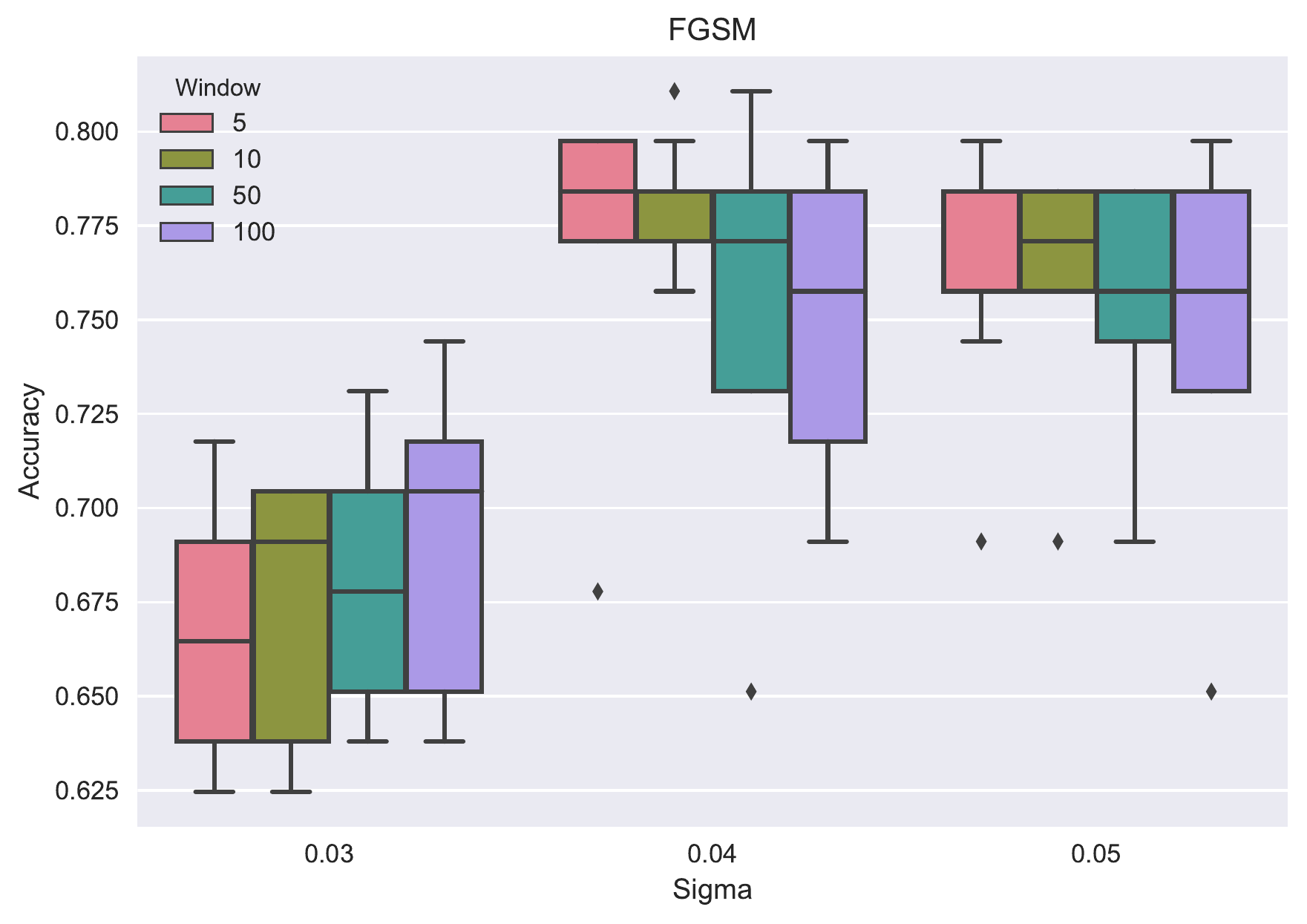}
   \includegraphics[width=0.49\linewidth]{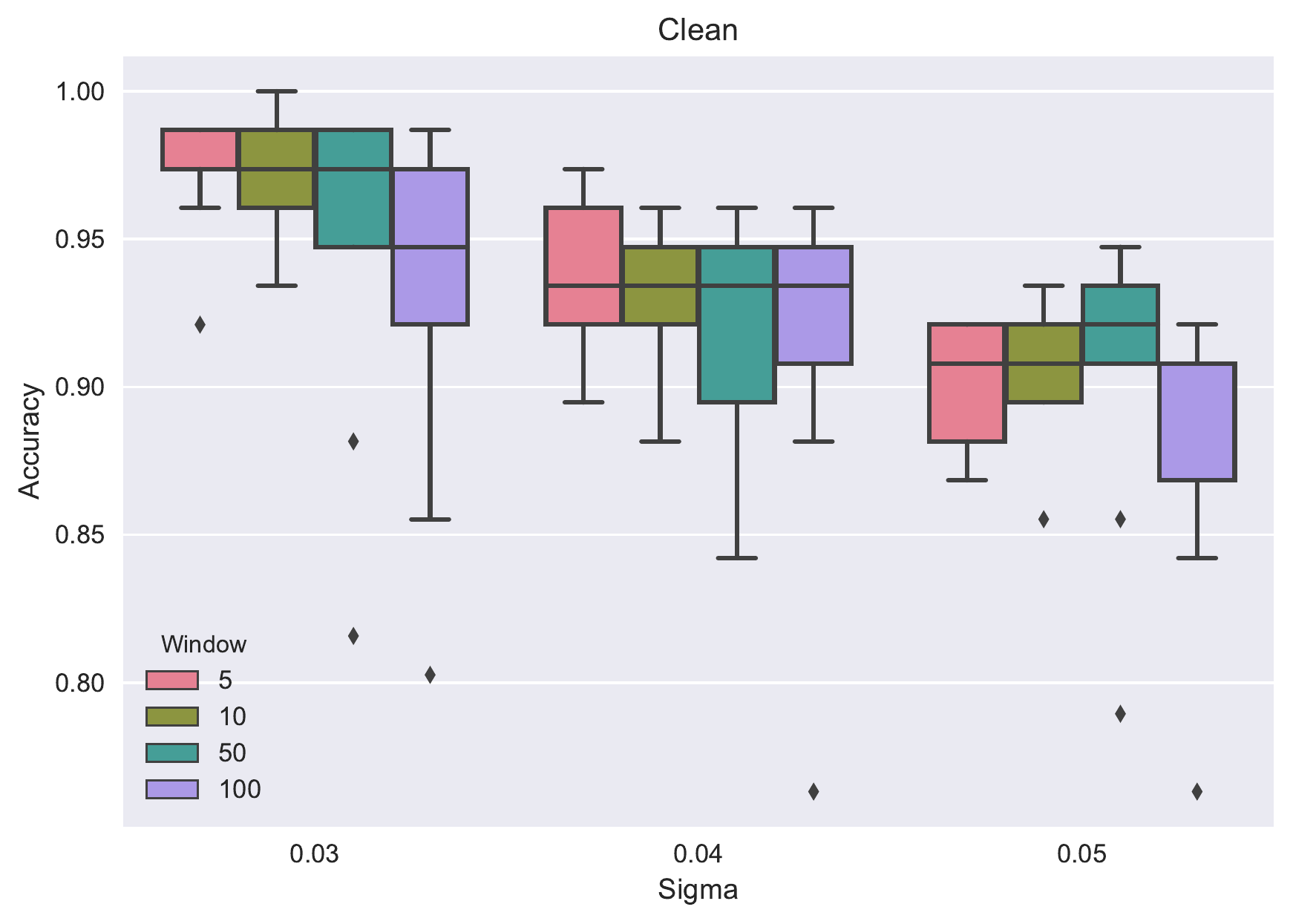}
   \caption{Linear search for model parameters}
   \label{fig:hyperparam}
\end{figure}

Our defense model has three hyper-parameters, which is significantly fewer than the classification models it seeks to protect, making it preferable over defenses which require retraining of the classifier such as ~\cite{Tramr2017TheSO,Meng2017MagNetAT}.
These three hyper-parameters are: $\sigma$, a coefficient for \textit{BayesShrink}, $r$, the window size for pixel deflection, and $K$, the number of pixel deflections to perform.
Using a reduced set of $300$ images from ImageNet Validation set, We perform a linear search over a small range of these hyper-parameters. 
These images are not part of the set used to show the results of our model.
A particular set of hyper-parameters may be optimal for one attack model, but not for another.  
This is primarily because attacks seek to minimize different $L_p$ norms, and therefore generate different types of noise.
To demonstrate the robustness of our defense, we select a single setting of the hyper-parameters to be used against all attack models.
Figure~\ref{fig:hyperparam} shows a visual indication of the variations in performance of each model across various hyper-parameter settings. 
In general, as the $K$ and $r$ increase, the variance of the resulting classification accuracy increases.  
This is primarily due to the stochastic nature of pixel deflection - as more deflections are performed over a wider window, a greater variety of transformed images can result.

\section{Results \& Discussion \label{sec:results}}

\begin{table}[h]
\small
\centering
\begin{tabular}{lcccc}
\textbf{Model} & \multicolumn{1}{l}{$|\boldsymbol{L}_2|$} & \multicolumn{1}{l}{\textbf{No Defense}} & \multicolumn{2}{l}{\textbf{With Defense}} \\ \hline
 & \multicolumn{1}{l}{} & \multicolumn{1}{l}{} & Single & Ens-10 \\ \cline{4-5} 
 \textbf{Clean} & 0.00 & 100 & 98.3 & \textbf{98.9} \\ \hline
\textbf{FGSM} & 0.05 & 20.0 & 79.9 & \textbf{81.5} \\
\textbf{IGSM} & 0.03 & 14.1 & 83.7 & \textbf{83.7} \\
\textbf{DFool} & 0.02 & 26.3 & 86.3 & \textbf{90.3} \\
\textbf{JSMA} & 0.02 & 25.5 & 91.5 & \textbf{97.0} \\
\textbf{LBFGS} & 0.02 & 12.1 & 88.0 & \textbf{91.6} \\
\textbf{C\&W} & 0.04 & 04.8 & 92.7 & \textbf{98.0}
\\
 \multicolumn{5}{c}{Large perturbations} \\ \hline
 \\
\textbf{FGSM} & 0.12 & 11.1 & 61.5 &  \textbf{70.4}\\
\textbf{IGSM} & 0.09 & 11.1 & 62.5 &  \textbf{72.5}\\
\textbf{DFool} & 0.08 & 08.0 & 82.4 &  \textbf{88.9}\\
\textbf{JSMA} & 0.05 & 22.1 & 88.9 &  \textbf{92.1}\\
\textbf{LBFGS} & 0.04 & 12.1 & 77.0 & \textbf{89.0} \\
\end{tabular}
\caption{ Params: $\sigma=0.04$, Window=10, Deflections=100 \\Top-1 accuracy on applying pixel deflection and wavelet denoising across various attack models. We evaluate non-efficient attacks at larger $|\boldmath{L}_P|$ which leave visible perturbations to show the robustness of our model.
 \label{tblresults}}
\end{table}
In Table~\ref{tblresults} we present results obtained by applying our transformation against various untargeted white-box attacks. 
Our method is agnostic to classifier architecture, and thus shows similar results across various classifiers. 
For brevity, we report only results on ResNet-50. 
Results for other classifiers are provided in Table~\ref{tblmodels}.
The accuracy on clean images without any defense is 100\% because we didn't test our defense on images which were misclassified before any attack.
We do not report results for targeted attacks as they are harder to generate~\cite{Carlini2017TowardsET} and easier to defend. 
Due to the stochastic nature of our model, we benefit from taking the majority prediction over ten runs; this is reported in Table~\ref{tblresults} as Ens-10.

We randomly sampled 10K images from ILSVRC2012 validation set; this contained all 1000 classes with minimum of $3$ images per class.

\begin{table}[h]
\small
\centering
\begin{tabular}{lcccc}
\textbf{Attack} & \multicolumn{1}{l}{$|\boldsymbol{L}_2|$} & \multicolumn{1}{l}{\textbf{No Defense}} & \multicolumn{2}{l}{\textbf{With Defense}} \\ \hline
 \multicolumn{3}{l}{Window=10, Deflections=100}  & Single & Ens-10 \\ \hline 
\small{Clean} & 0.00 & 100 & 98.1 & \textbf{98.9} \\ 
\small{FGSM} & 0.04 & 19.2 & 79.7 & \textbf{81.2} \\
\small{IGSM} & 0.03 & 11.8 & 81.7 & \textbf{82.4} \\
\small{DFool} & 0.02 & 18.0 & 87.7 & \textbf{92.4} \\
\small{JSMA} & 0.02 & 24.9 & 93.0 & \textbf{98.1} \\
\small{LBFGS} & 0.02 & 11.6 & 90.3 & \textbf{93.6} \\
\small{C\&W} & 0.04 & 05.2 & 93.1 & \textbf{98.3}   \\ \hline
\end{tabular}
\caption{Top-1 accuracy of our model on various attack models.}
\end{table}

\subsection{Results on various classifiers}
Original classification accuracy of each classifier on selected $1000$ images is reported in the table.
However, we omit the images that were originally incorrectly classified, thus the accuracy of clean images without defense is always $100\%$.
Weights for each classifier were obtained from Tensorflow GitHub repository
\footnote{https://github.com/tensorflow/models/tree/master/research/slim\#Pretrained}. \\

\begin{table}[h]
\small
\centering
\begin{tabular}{lcccc}
\textbf{Model} & \multicolumn{1}{l}{$|\boldsymbol{L}_2|$} & \multicolumn{1}{l}{\textbf{No Defense}} & \multicolumn{2}{l}{\textbf{With Defense}} \\ \hline
 & \multicolumn{1}{l}{} & \multicolumn{1}{l}{} & Single & Ens-10 \\ \cline{4-5} 
 \\
 \multicolumn{5}{c}{ResNet-50, original classification $76\%$} \\ \hline
 \\
\textbf{Clean} & 0.00 & 100 & 98.3 & \textbf{98.9} \\ \hline
\textbf{FGSM} & 0.05 & 20.0 & 79.9 & \textbf{81.5} \\
\textbf{IGSM} & 0.03 & 14.1 & 83.7 & \textbf{83.7} \\
\textbf{DFool} & 0.02 & 26.3 & 86.3 & \textbf{90.3} \\
\textbf{JSMA} & 0.02 & 25.5 & 91.5 & \textbf{97.0} \\
\textbf{LBFGS} & 0.02 & 12.1 & 88.0 & \textbf{91.6} \\
\textbf{C\&W} & 0.04 & 04.8 & 92.7 & \textbf{98.0} \\
 \\
 \multicolumn{5}{c}{VGG-19, original classification $71\%$} \\ \hline
 \\
\textbf{Clean} & 0.00 & 100 & 99.8 & \textbf{99.8} \\ \hline
\textbf{FGSM} & 0.05 & 12.2 & 79.3 & \textbf{81.3} \\
\textbf{IGSM} & 0.04 & 9.79 & 79.2 & \textbf{81.6} \\
\textbf{DFool} & 0.01 & 23.7 & 83.9 & \textbf{91.6} \\
\textbf{JSMA} & 0.01 & 29.1 & 95.8 & \textbf{98.5} \\
\textbf{LBFGS} & 0.03 & 13.8 & 83.0 & \textbf{93.9} \\
\textbf{C\&W} & 0.04 & 0.00 & 93.1 & \textbf{97.6} \\
 \\
 \multicolumn{5}{c}{Inception-v3, original classification $78\%$} \\ \hline
 \\
\textbf{Clean} & 0.00 & 100 & 98.1 & \textbf{98.5} \\ \hline
\textbf{FGSM} & 0.05 & 22.1 & 85.8 & \textbf{87.1} \\
\textbf{IGSM} & 0.04 & 15.5 & \textbf{89.7} & 89.1 \\
\textbf{DFool} & 0.02 & 27.2 & 82.6 & \textbf{85.3} \\
\textbf{JSMA} & 0.02 & 24.2 & 93.7 & \textbf{98.6} \\
\textbf{LBFGS} & 0.02 & 12.5 & 87.1 & \textbf{91.0} \\
\textbf{C\&W} & 0.04 & 07.1 & 93.9 & \textbf{98.5} \\
\end{tabular}
\caption{Params: $\sigma=0.04$, Window=10, Deflections=100 \\Top-1 accuracy on applying pixel deflection and wavelet denoising across various attack models. 
 \label{tblmodels}}
\end{table}

\subsection{Comparison of results} - 
There are two main challenges when seeking to compare defense models. 
First, many attack and defense techniques primarily work on smaller images, such as those from CIFAR and MNIST.
The few proposed transformation based defense techniques which work on larger-scale images are extremely recent, and currently under review~\cite{MitigatingAnon208,CounteringAIGuo17}.
Second, because different authors target both different $|\boldmath{L}_P|$ norms and different perturbation magnitudes, it is difficult to balance the strength of various attacks.
We achieved $98\%$ recovery on C\&W with $|\boldmath{L}_2|$ of 0.04 on ResNet-50, where Xie \etal~\cite{MitigatingAnon208} reports $97.1\%$ on ResNet-101 and $98.8\%$ on ens-adv-Inception-ResNet-v2.
ResNet-101 is as stronger classifier than ResNet-50 and ens-adv-Inception-Resnet-v2~\cite{Tramr2017EnsembleAT} is an ensemble of classifiers specifically trained with adversarial augmentation. 
They do not report the $|\boldmath{L}_2|$ norm of the adversarial perturbations, and predictions are made on an ensemble of 21 crops.
Guo \etal~\cite{CounteringAIGuo17} have reported (normalized) accuracy of $92.1\%$ on C\&W with $|\boldmath{L}_2|$ of 0.06, and their predictions are on an ensemble of 10 crops.

To present a fair comparison across various defenses we only measure the fraction of images which are no longer misclassified after the transformation.  
This ratio is known as  Destruction Rate and was originally proposed in [24]. 
Value of $1$ means all the misclassified images due to the adversary are correctly classified after the transformation.

\begin{table}[h]
\small
\centering
\setlength{\tabcolsep}{0.4em}
\begin{tabular}{lllll}
\textbf{Defense}              & \textbf{FGSM}  & \textbf{IGSM}  & \textbf{DFool} & \textbf{C\&W}  \\ \hline
\multicolumn{5}{c}{Feature Squeezing (Xu et al \cite{FeatureSqueezingXu2017})}                                                \\ \hline
\small{(a) Bit Depth (2 bit)} & 0.132          & 0.511          & 0.286          & 0.170          \\
\small{(b) Bit Depth (5 bit)} & 0.057          & 0.022          & 0.310          & 0.957          \\
\small{(c) Median Smoothing (2x2)} & 0.358          & 0.422          & 0.714          & 0.894          \\
\small{(d) Median Smoothing (3x3)} & 0.264          & 0.444          & 0.500          & 0.723          \\
\small{(e) Non-local Mean (11-3-2)} & 0.113          & 0.156          & 0.357          & 0.936          \\
\small{(f) Non-local Mean (13-3-4)} & 0.226          & 0.444          & 0.548          & 0.936          \\
\small{Best model (b) + (c) + (f)} & 0.434          & 0.644          & 0.786          & 0.915          \\ \hline
\multicolumn{5}{c}{Random resizing + padding (Xie \etal \cite{MitigatingAnon208} )}                               \\ \hline
\small{Pixel padding}    & 0.050          & -              & 0.972          & 0.698          \\
\small{Pixel resizing}    & 0.360          & -              & 0.974          &    0.971       \\
\small{Padding + Resizing}   & 0.478          & -              & \textbf{0.983} & 0.969          \\ \hline
\multicolumn{5}{c}{Quilting + TVM (Guo \etal \cite{CounteringAIGuo17} )}                                         \\ \hline
\small{Quilting}        & 0.611          & 0.862          & 0.858          & 0.843          \\
\small{TVM + Quilting}& 0.619          & 0.866          & 0.866          & 0.841          \\
\small{Cropping + TVM + Quilting} & 0.629          & 0.882          & 0.883          & 0.859          \\ \hline
\multicolumn{5}{c}{Our work: PD - Pixel Deflection, R-CAM: Robust CAM}                                                                    \\ \hline
\small{PD}  & 0.735          & 0.880          & 0.914          & 0.931          \\
\small{PD + R-CAM}       & 0.746          & 0.912          & 0.911          & 0.952          \\
\small{PD + R-CAM + DCT}       & 0.737          & 0.906          & 0.874         & 0.930          \\
\small{PD + R-CAM + DWT}  & \textbf{0.769} & \textbf{0.927} & 0.948          & \textbf{0.981} \\ \hline
\end{tabular}
\caption{Destruction Rate of various defense techniques. $|L_2|$ lies between $0.02 - 0.06$ and classifier accuracy is $76\%$.
We only include the Black-box attacks, where the attack model is not aware of the defense techniques. Single Pattern Attack and Ensemble pattern attack as reported in Xie et al \cite{MitigatingAnon208} are not reported.}
\end{table}

\begin{figure}[h]
\centering
   \includegraphics[width=0.49\linewidth]{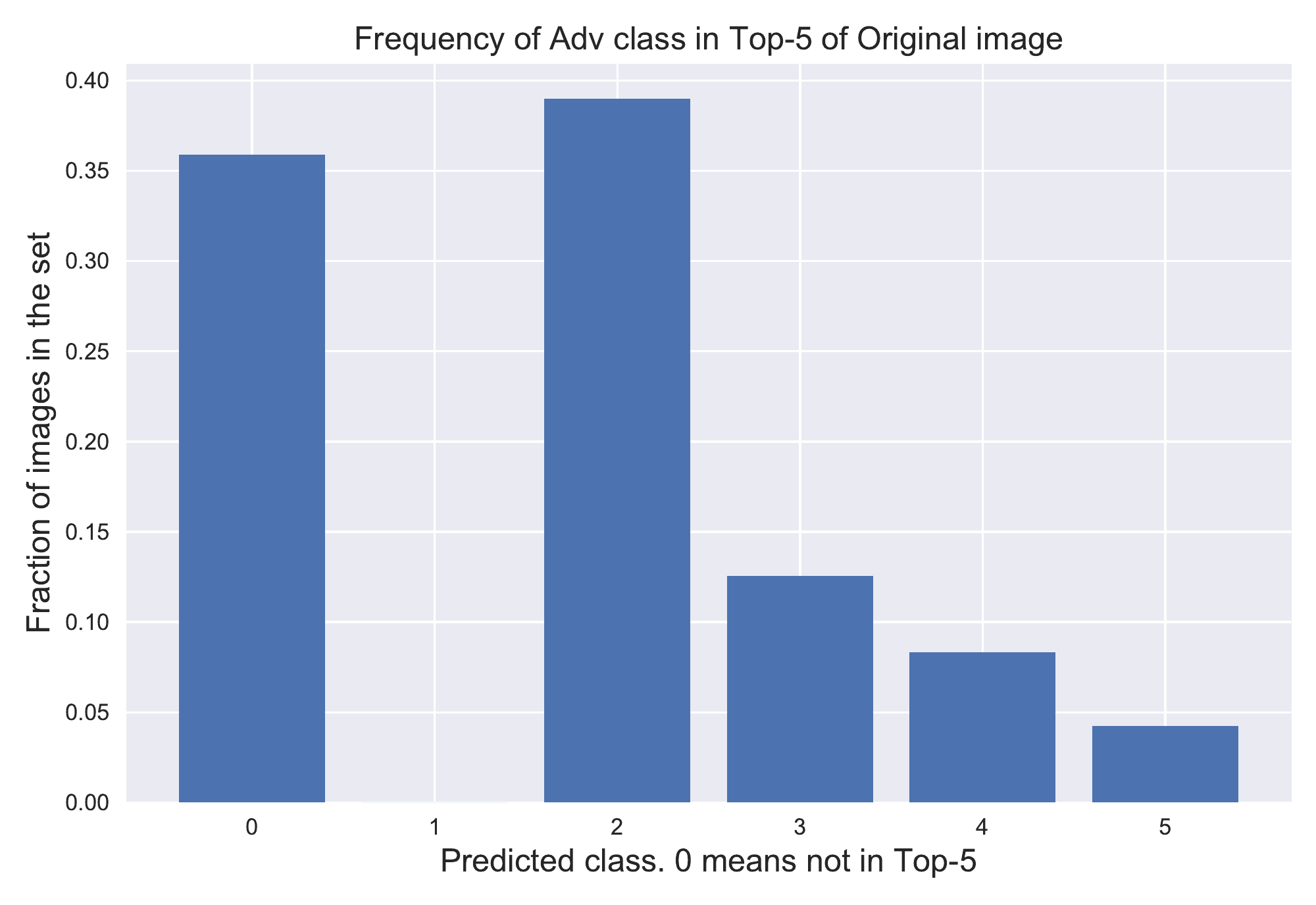}
   \includegraphics[width=0.49\linewidth]{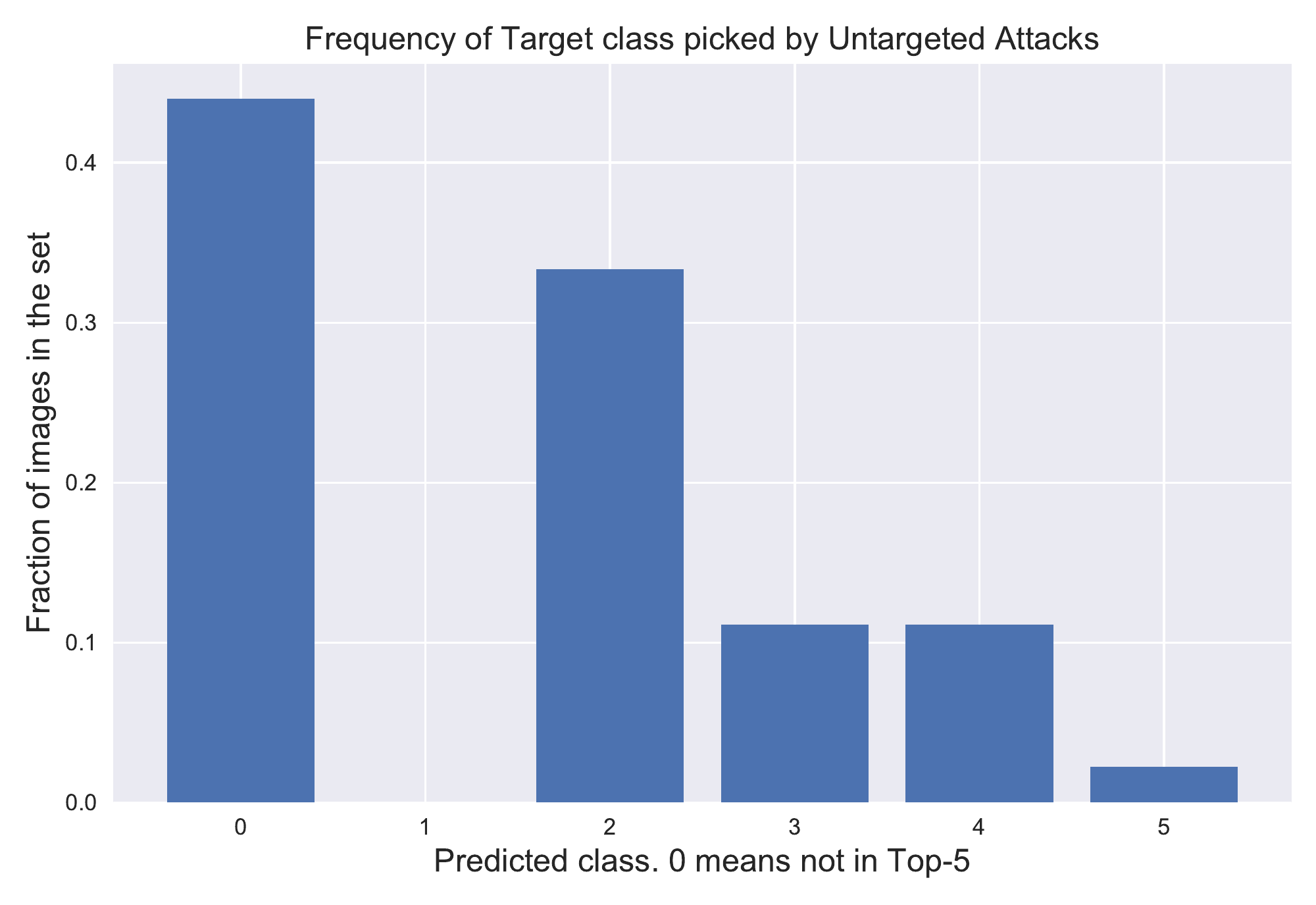}
   \caption{Left: Rank of adversarial class within the top-5 predictions for original images.  Right: Rank of original class within the top-5 predictions for adversarial images.  In both cases, 0 means the class was not in the top-5.}
   \label{fig:untarget}
\end{figure}
As seen in Figure~\ref{fig:untarget}, the predicted class of the perturbed image is very frequently among the classifier's top-5 predictions for the original image.
In fact, nearly $40\%$ of the time, the adversarial class was the second most-probable class of the original image.
Similarly, the original classification will often remain in the top-5 predictions for the adversarial image.
Unlike Kurakin \etal~\cite{Kurakin2016AdversarialEI}, our results are in terms of top-1 accuracy, as this matches the objective of the attacker.
While top-1 accuracy is a more lenient metric for an attack method (due to the availability of nearly-synonymous alternatives to most classes in ImageNet-1000), it is a more difficult metric for a defense, as we must exactly recover the correct classification.
These facts render top-5 accuracy an unsuitable metric for measuring the efficacy of a defense.
Results reported for Carlini \& Wagner~\cite{Carlini2017TowardsET} attacks are only for $L_2$ loss, even though they can be applied for $L_0$ and $L_\infty$. 
Carlini \& Wagner attack has been shown to be effective with MNIST and CIFAR but their efficacy against large images is limited due to expensive computation. 

\subsection{Ablation studies}

Previous work~\cite{Kurakin2016AdversarialEI,Dziugaite2016ASO} has demonstrated the efficacy of JPEG compression as a defense against adversarial attacks due to its denoising properties.
Das \etal~\cite{Das2017KeepingTB} demonstrate that increasing the severity of JPEG compression defeats a larger percentage of attacks, but at the cost of accuracy on clean image.
As our method employs a conceptually similar method to reduce adversarial noise via thresholding in a wavelet domain, we use JPEG as a baseline for comparison.
%
In Table~\ref{tblimpact}, we report accuracy with and without wavelet denoising with soft thresholding.
While JPEG alone is effective against only a few attacks, the combination of JPEG and pixel deflection performs better than pixel deflection alone.
The best results are obtained from pixel deflection and wavelet denoising.
Adding JPEG on top of these leads to a drop in performance. 
\begin{table}[h]
\small
\centering
\begin{tabular}{lllllll}
\textbf{Model} & \textbf{JPG} & \textbf{WD} & \textbf{PD} & \textbf{\begin{tabular}[c]{@{}l@{}}\,PD\\ JPG\end{tabular}} & \textbf{\begin{tabular}[c]{@{}l@{}}WD\\ \,PD\\ JPG\end{tabular}} & \textbf{\begin{tabular}[c]{@{}l@{}}WD\\ \,PD\end{tabular}} \\ \hline
\textbf{Clean} &  96.1 &  98.7 &  97.4 &      96.1 &           96.1 &     \textbf{98.9} \\
\textbf{FGSM} &  49.1 &  40.6 &  79.7 &      81.1 &           78.8 &     \textbf{81.5} \\
\textbf{IGSM} &  49.1 &  31.2 &  82.4 &      82.4 &           79.7 &     \textbf{83.7} \\
\textbf{DFool} &  67.8 &  61.1 &  86.3 &      86.3 &           86.3 &     \textbf{90.3} \\
\textbf{JSMA} &  91.6 &  89.1 &  95.7 &      93.0 &           93.0 &     \textbf{97.0} \\
\textbf{LBFGS} &  71.8 &  67.2 &  90.3 &      89.1 &           88.9 &     \textbf{91.6} \\
\textbf{C\&W} &  85.5 &  95.4 &  95.4 &      94.1 &           93.4 &     \textbf{98.0} \\
\end{tabular}
\caption{ Params: $\sigma=0.04$, Window=10, Deflections=100 \\
Ablation study of pixel deflection (PD) in combination with wavelet denoising (WD) and JPEG compression.\label{tblimpact}}
\end{table}

\begin{table}[h]
\small
\centering
\begin{tabular}{lrrrr}
\textbf{Model} & \multicolumn{1}{c}{\textbf{\begin{tabular}[c]{@{}c@{}}Hard\end{tabular}}} &  \multicolumn{1}{c}{\textbf{\begin{tabular}[c]{@{}c@{}}VISU\end{tabular}}} & \multicolumn{1}{c}{\textbf{\begin{tabular}[c]{@{}c@{}}SURE\end{tabular}}} & \multicolumn{1}{c}{\textbf{\begin{tabular}[c]{@{}c@{}}Bayes\end{tabular}}} \\ \hline
 \textbf{Clean} &               39.5 &        96.1 &        92.1 &         \textbf{98.9} \\
  \textbf{FGSM} &               35.9 &        63.8 &        79.7 &         \textbf{81.5} \\
  \textbf{IGSM} &               42.5 &        67.8 &        81.1 &         \textbf{83.7} \\
 \textbf{DFool} &               37.2 &        78.4 &        87.7 &         \textbf{90.3} \\
 \textbf{JSMA} &               39.9 &        93.0 &        93.0 &         \textbf{97.0} \\
  \textbf{LBFGS} &               37.2 &        81.1 &        90.4 &         \textbf{91.6} \\
   \textbf{C\&W} &               36.8 &        93.4 &        92.8 &         \textbf{98.0} \\
\end{tabular}
\caption{ Params: $\sigma=0.04$, Window=10, Deflections=100 \\
Comparison of various thresholding techniques, after application of pixel deflection. \label{tblshrink}}
\end{table}

In Table~\ref{tblshrink} we present a comparison of various shrinkage methods on wavelet coefficients after pixel deflection.
For the impact of coefficient thresholding in the absence of pixel deflection, see Table~\ref{tblimpact}.
BayesShrink, which learns separate Gaussian parameters for each coefficient, does better than other soft-thresholding techniques.
A brief overview of these shrinkage techniques are provided in Section~\ref{sec:contentthresholding}, for more thorough review on BayesShrink, VisuShrink and SUREShrink we refer the reader to~\cite{BayesShrinkChang2000} ~\cite{VISUDonoho1994IdealSA} and ~\cite{SUREDonoho1992AdaptingTU} respectively. 
VisuShrink is a faster technique as it uses a universal threshold but that limits its applicability on some images. 
SUREShrink has been shown to perform well with compression but as evident, in our results, it is less well suited to denoising.





\begin{table}[h]
\small
\centering
\begin{tabular}{llcllll}
\multicolumn{1}{c}{\textbf{Attack}} & \multicolumn{1}{c}{\textbf{$|L_2$}} & \multicolumn{1}{c}{\textbf{No Defense}} & \multicolumn{4}{c}{\textbf{\begin{tabular}[c]{@{}c@{}}With Defense\end{tabular}}} \\ \hline
                                   \multicolumn{3}{r}{Window=10, Deflections $\longrightarrow$}                               & \textbf{10}                    & \textbf{100}                    & \textbf{1K}                    & \textbf{10K}                    \\ \hline
\textbf{Clean}                              & 0.00                            &  100                                    & \textbf{98.4}                  & 98.1                  & 94.7                    & 80.3                     \\
\textbf{FGSM}                               & 0.04                            &  19.2                                   & 75.7                  & \textbf{79.7}                  & 71.7                    & 69.1                     \\
\textbf{IGSM}                               & 0.03                            &  13.8                                   & 78.4                  &\textbf{81.7}                & 75.2                    & 71.2                     \\
\textbf{DFool}                              & 0.02                            &  25.0                                   & 83.7                  & \textbf{87.7}                & 81.0                    & 77.0                     \\
\textbf{JSMA}                               & 0.02                            &  25.9                                   & 91.7                  &\textbf{93.0}                 & 87.7                    & 67.7                     \\
\textbf{LBFGS}                              & 0.02                            &  11.6                                   & 85.0                  & \textbf{90.3}               & 82.4                    & 73.0                     \\
\textbf{C\&W}                               & 0.04                            &  05.2                                   & 89.4                  &\textbf{93.1}                & 86.8                    & 69.7                    \\ \hline
\end{tabular}
\vspace{-3mm}
\caption{Top-1 accuracy with different deflections.}
\vspace{-3mm}
\end{table}

\begin{table}[h]
\small
\centering
\begin{tabular}{lllllll}
\multicolumn{1}{c}{\textbf{Attack}} & \multicolumn{1}{c}{\textbf{L2}} & \multicolumn{1}{c}{\textbf{No Defense}} & \multicolumn{4}{c}{\textbf{\begin{tabular}[c]{@{}c@{}}With Defense\end{tabular}}} \\ \hline
                                   \multicolumn{3}{r}{Deflections=100, Window $\longrightarrow$}                                         & \textbf{5}                     & \textbf{10}                     & \textbf{50}                      & \textbf{100}                     \\ \hline
\textbf{Clean}                              & 0.00                            &  100                                    &\textbf{98.6}            & 98.1                   & 96.4                    & 94.4                     \\
\textbf{FGSM}                               & 0.04                            &  19.2                                   & \textbf{79.7}                & \textbf{79.7}                 & 78.4                    & 76.7                     \\
\textbf{IGSM}                               & 0.03                            &  13.8                                   & 81.0                  & \textbf{81.7}                 & 79.7                    & 78.4                     \\
\textbf{DFool}                              & 0.02                            &  25.0                                   & 86.4                  &\textbf{87.7}                  & 87.7                    & 85.0                     \\
\textbf{JSMA}                               & 0.02                            &  25.9                                   & 92.3                  & \textbf{93.0}                 & 91.7                    & 90.3                     \\
\textbf{LBFGS}                              & 0.02                            &  11.6                                   & 89.4                  & \textbf{90.3}                  & 89.0                    & 88.1                     \\
\textbf{C\&W}                               & 0.04                            &  05.2                                   & 91.8                  & \textbf{93.1}                  & 90.5                    & 89.2                      \\ \hline
\end{tabular}
\caption{Top-1 accuracy with different window sizes.}
\end{table}


\begin{table}[h]
\small
\centering
\begin{tabular}{lllll} \hline
\multicolumn{5}{c}{Sampling technique (Random Pixel)} \\ \hline
Window $\longrightarrow$   & \textbf{5} & \textbf{10} & \textbf{50} & \textbf{100} \\ \hline
Uniform  & \textbf{86.7}          & \textbf{87.5}          & \textbf{86.1}           & \textbf{84.6}           \\
Gaussian & 80.0          & 81.4          & 79.0           & 76.4          \\ \hline
\multicolumn{5}{c}{Replacement technique (Uniform Sampling)} \\ \hline
Window $\longrightarrow$   & \textbf{5} & \textbf{10} & \textbf{50} & \textbf{100} \\ \hline
Min & 73.0          & 64.4          & 49.1           & 44.3          \\
Max & 69.7          & 63.8          & 51.9           & 45.4          \\ 
Mean & 83.6         & 72.3         & 57.2           & 49.1          \\
Random & \textbf{86.7}       & \textbf{87.5}         & \textbf{86.1}           & \textbf{84.6}         \\ \hline
\multicolumn{5}{c}{Various Denoising Techniques} \\ \hline
\textbf{Bilateral} & \textbf{Anisotropic} & \textbf{TVM} & \textbf{Deconv} & \textbf{Wavelet} \\ \hline
78.1     & 84.1        & 77.26           & 85.12  & \textbf{87.5  } 

\end{tabular}
\caption{Top-1 accuracy averaged across all six attacks.}
\end{table}


\section{Conclusion}
Motivated by the robustness of CNNs and the fragility of adversarial attacks, we have presented a technique which combines a computationally-efficient image transform, \emph{pixel deflection}, with soft wavelet denoising. 
This combination provides an effective defense against state-of-the-art adversarial attacks. 
We show that most attacks are agnostic to semantic content, and using \emph{pixel deflection} with probability inversely proportionate to robust activation maps (R-CAM) protects regions of interest.
In ongoing work, we seek to improve our technique by adapting hyperparameters based on the features of individual images. Additionally, we seek to integrate our robust activation maps with wavelet denoising.


\section{Acknowledgement}
We would like to thank NVIDIA for donating the GPUs used for this research.
We would also like to thank Ryan Marcus, Brandeis University, for reviewing the paper.

\clearpage
\onecolumn


\clearpage
\begin{figure}[h]
   \includegraphics[width=1\linewidth]{figures/robust_cam.pdf}
   \includegraphics[width=1\linewidth]{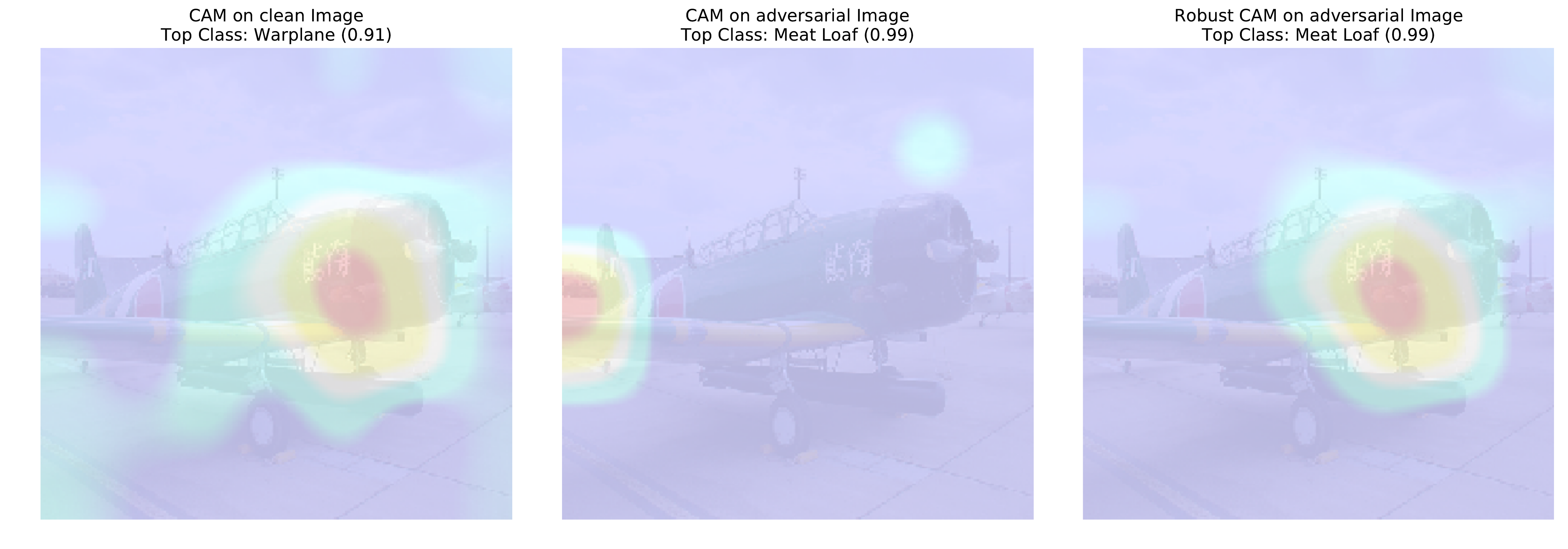}
   \includegraphics[width=1\linewidth]{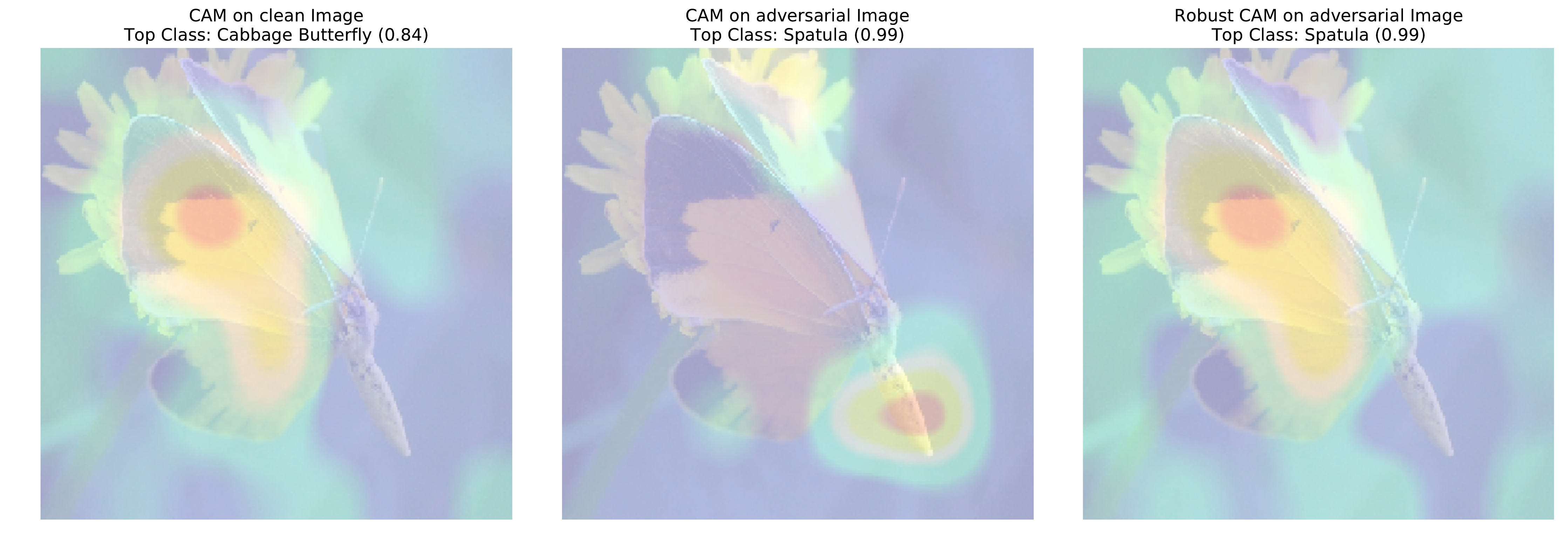}
   \includegraphics[width=1\linewidth]{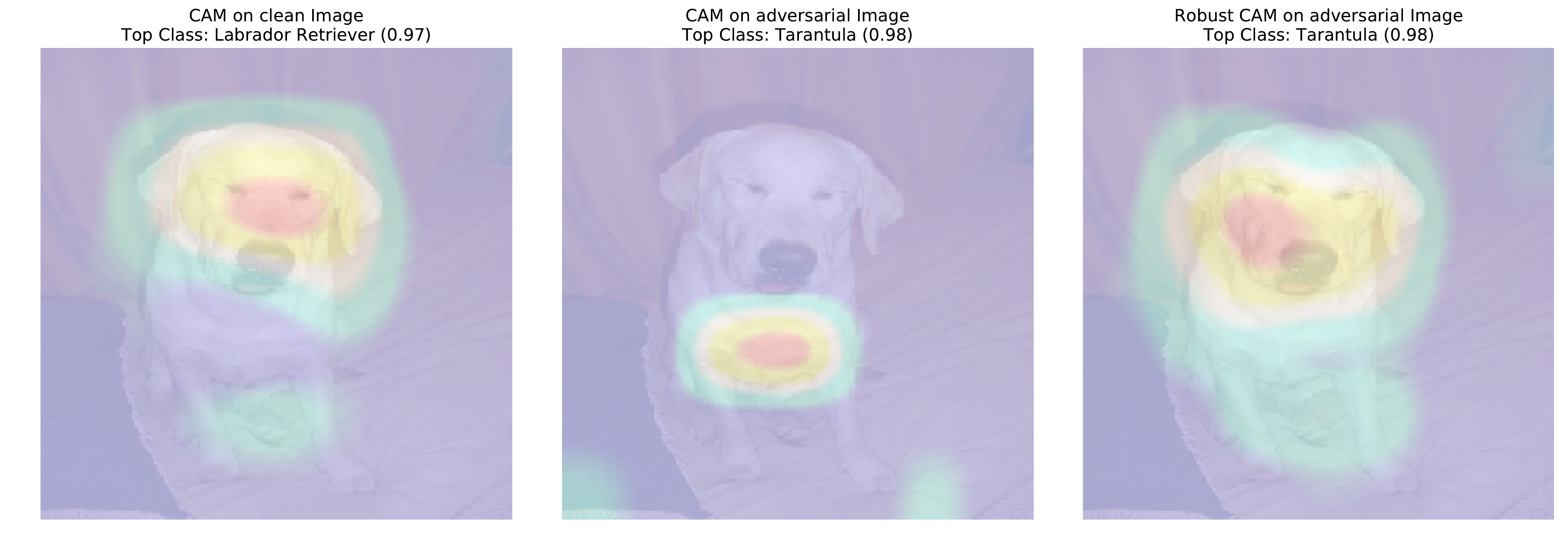}
   \caption{Comparison of Class activation maps and Robust Activation maps}
\end{figure}

\clearpage
Full size figures of Figure 5 (Linear search for model parameters on training data)
\begin{figure}[h]
   \includegraphics[width=1\linewidth]{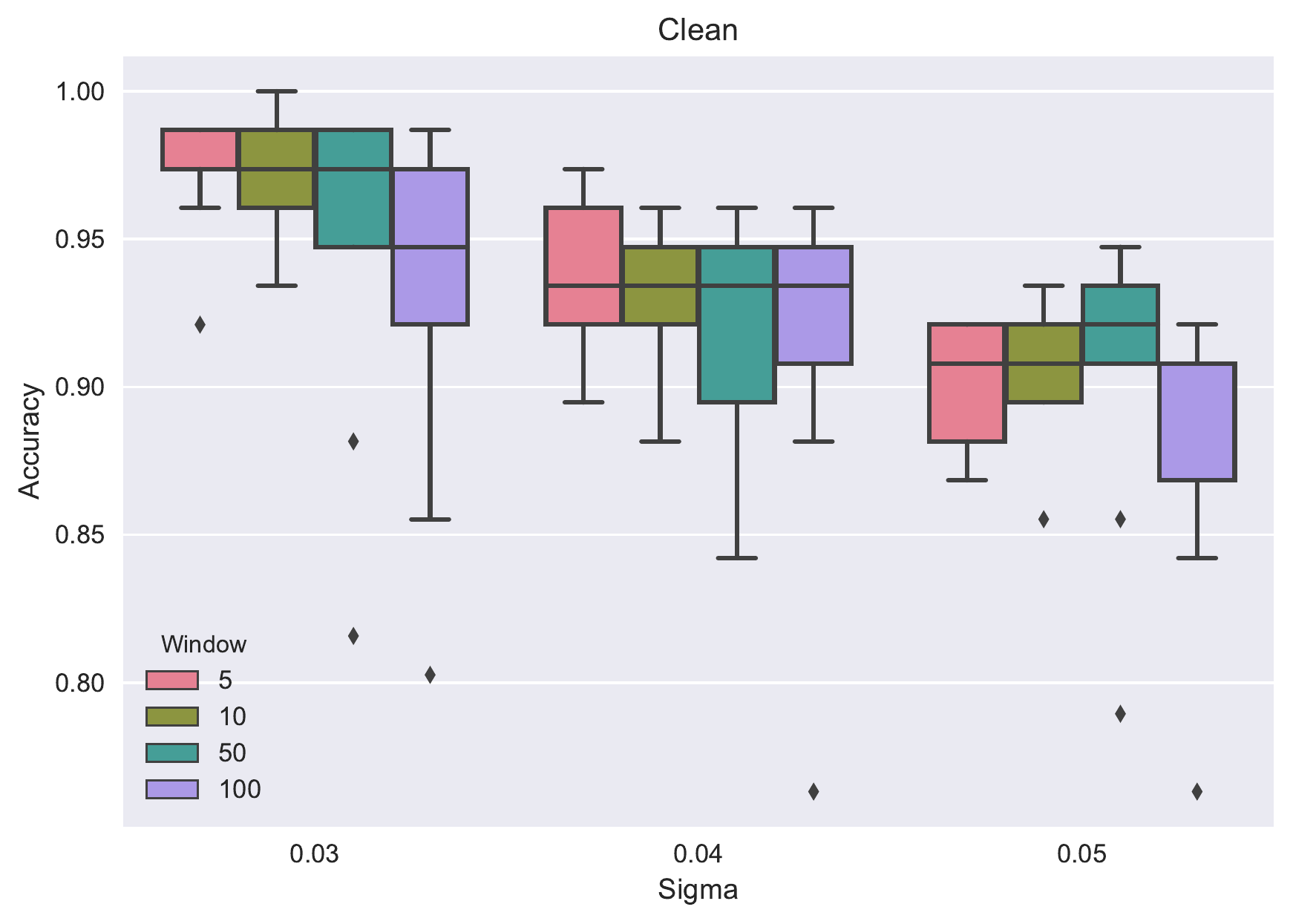}
\end{figure}
\begin{figure}[h]
   \includegraphics[width=1\linewidth]{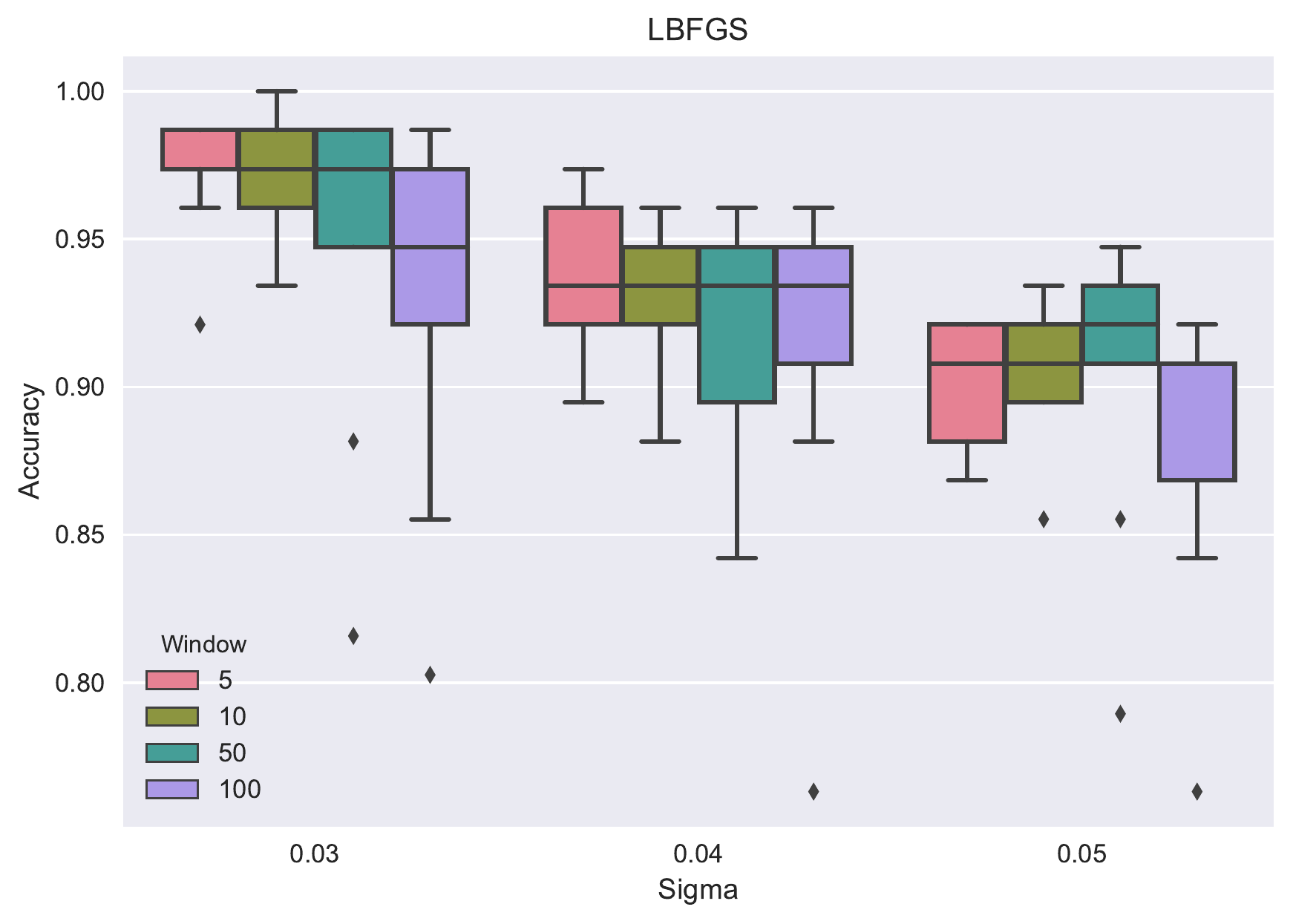}
   \includegraphics[width=1\linewidth]{figures/CW.pdf}
\end{figure}
\begin{figure}[h]
   \includegraphics[width=1\linewidth]{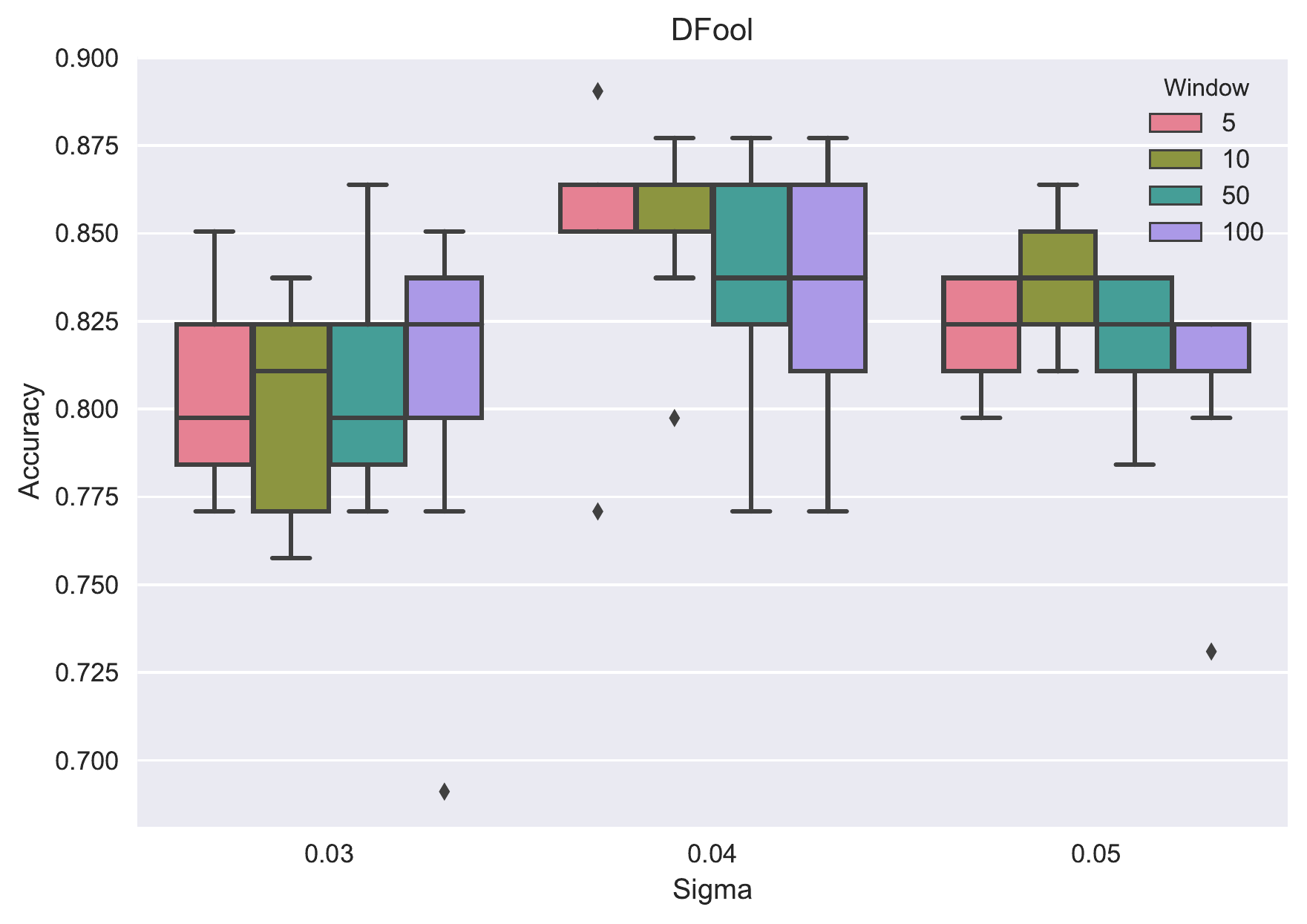}
   \includegraphics[width=1\linewidth]{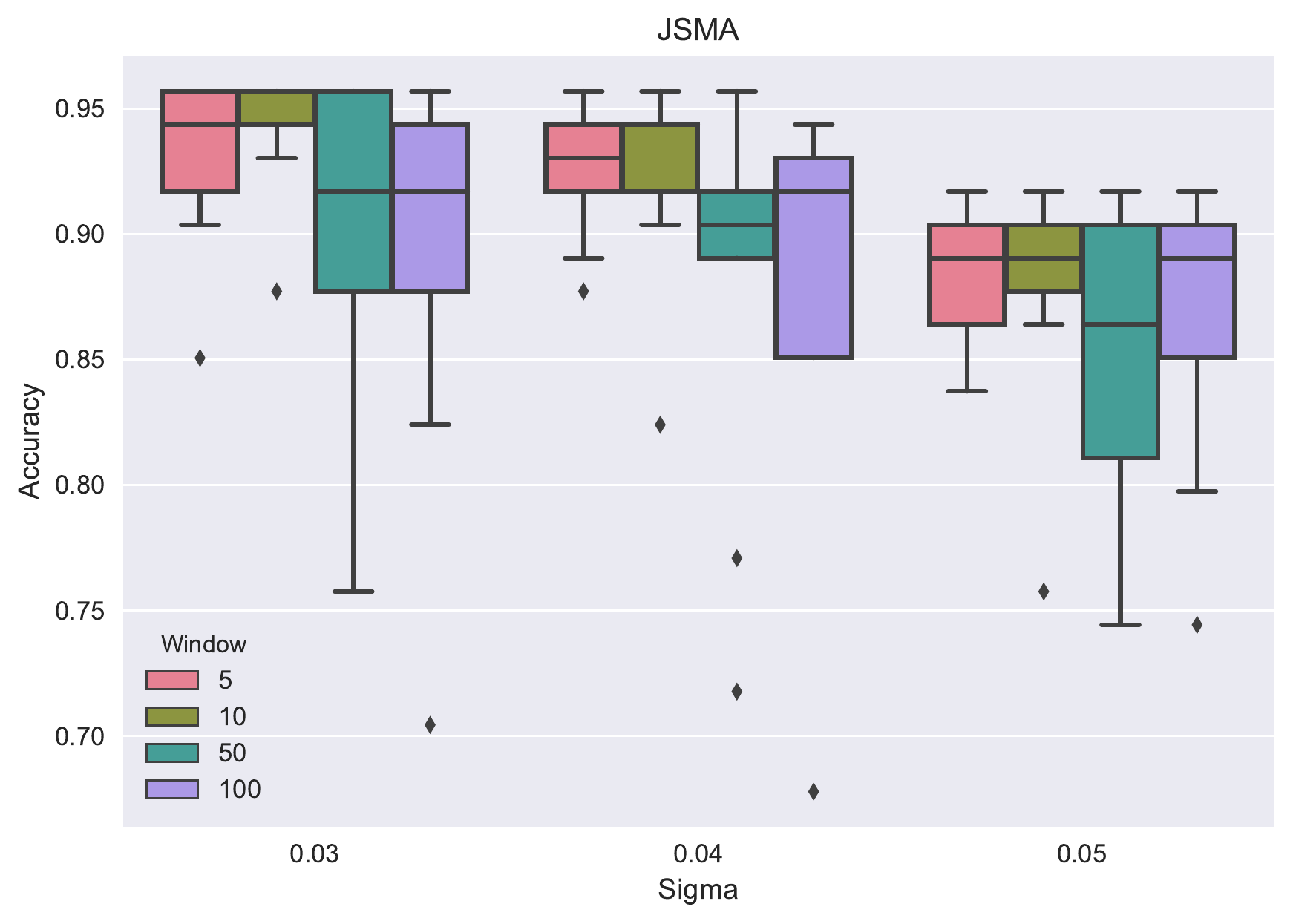}
\end{figure}
\begin{figure}[h]
   \includegraphics[width=1\linewidth]{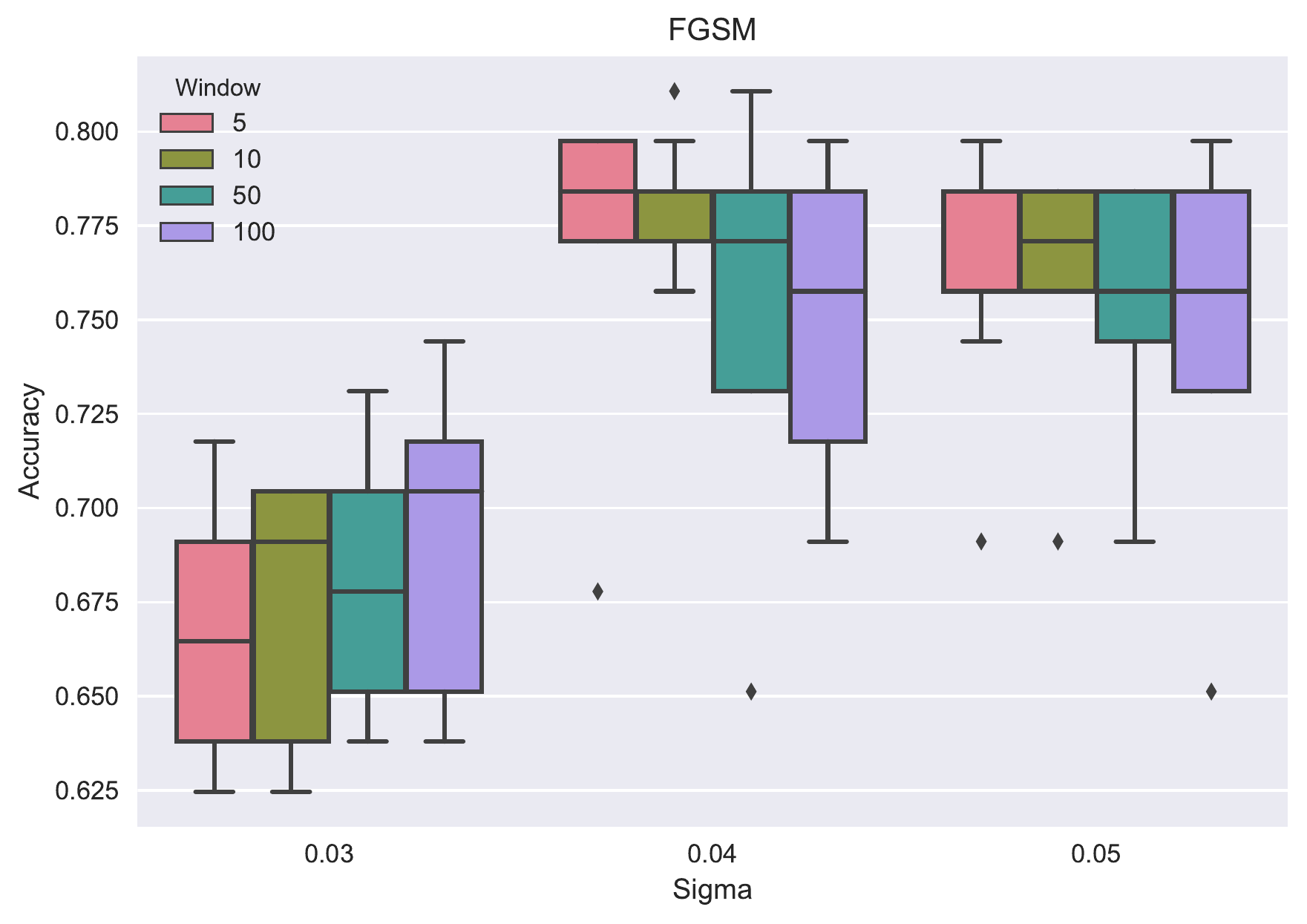}
   \includegraphics[width=1\linewidth]{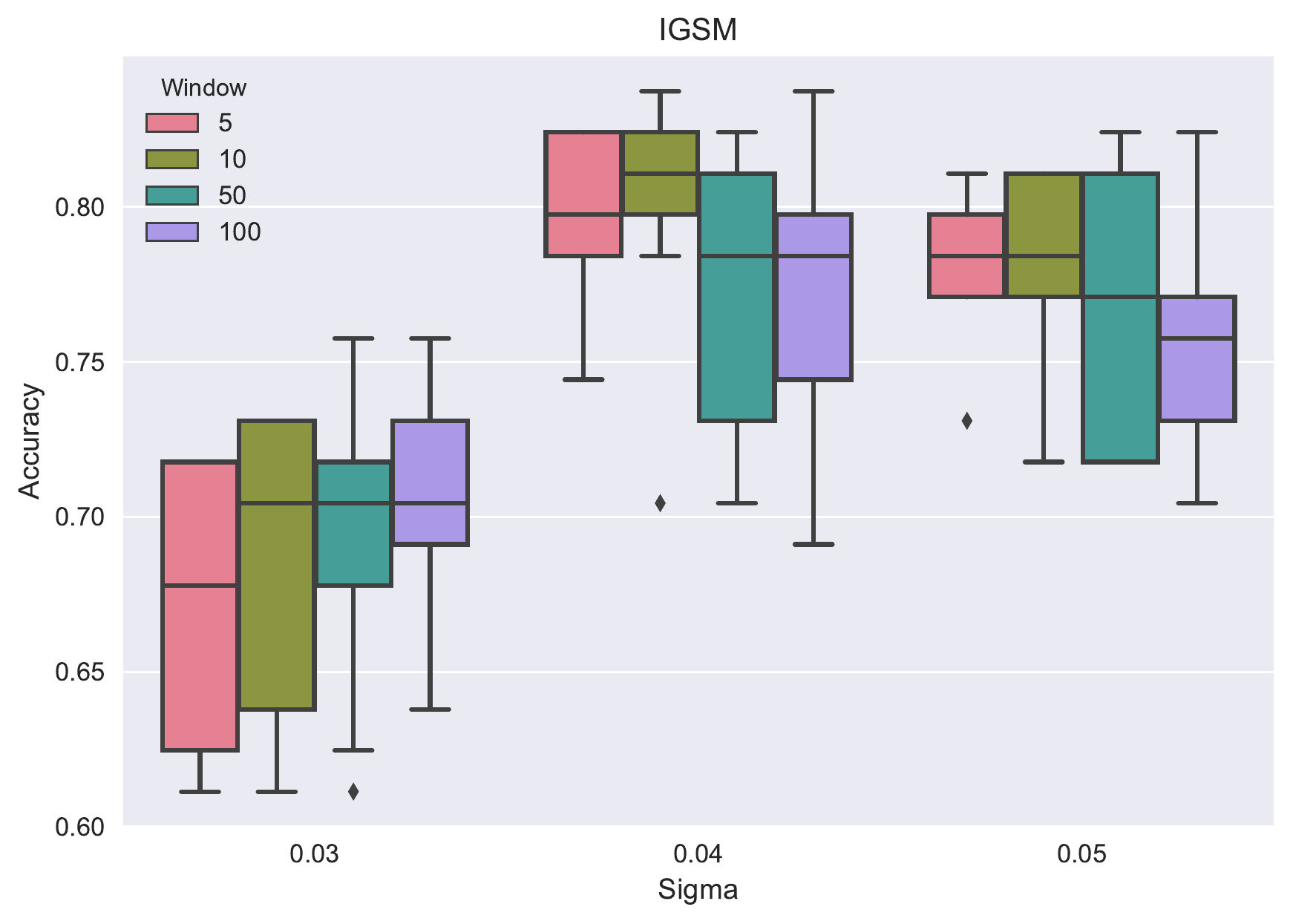}
\end{figure}

\clearpage
{\small
\bibliographystyle{ieee}
\bibliography{egbib}

\begin{thebibliography}{10}\itemsep=-1pt

\bibitem{JPEG2000Adams2001}
M.~D. Adams.
\newblock The jpeg - 2000 still image compression standard � ( last revised :
  June 30 , 2001 ).
\newblock 2001.

\bibitem{akhtar2017defense}
N.~Akhtar, J.~Liu, and A.~Mian.
\newblock Defense against universal adversarial perturbations.
\newblock {\em arXiv preprint arXiv:1711.05929}, 2017.

\bibitem{DWTantonini1992image}
M.~Antonini, M.~Barlaud, P.~Mathieu, and I.~Daubechies.
\newblock Image coding using wavelet transform.
\newblock {\em IEEE Transactions on image processing}, 1(2):205--220, 1992.

\bibitem{Bottou1997GlobalTO}
L.~Bottou, Y.~Bengio, and Y.~LeCun.
\newblock Global training of document processing systems using graph
  transformer networks.
\newblock In {\em CVPR}, 1997.

\bibitem{EasilyDetectedCarlini2017}
N.~Carlini and D.~A. Wagner.
\newblock Adversarial examples are not easily detected: Bypassing ten detection
  methods.
\newblock In {\em AISec@CCS}, 2017.

\bibitem{Carlini2017TowardsET}
N.~Carlini and D.~A. Wagner.
\newblock Towards evaluating the robustness of neural networks.
\newblock {\em 2017 IEEE Symposium on Security and Privacy (SP)}, 2017.

\bibitem{BayesShrinkChang2000}
S.~G. Chang, B.~Yu, and M.~Vetterli.
\newblock Adaptive wavelet thresholding for image denoising and compression.
\newblock {\em IEEE transactions on image processing : a publication of the
  IEEE Signal Processing Society}, 9 9:1532--46, 2000.

\bibitem{Chattopadhyay2017GradCAMGG}
A.~Chattopadhyay, A.~Sarkar, P.~Howlader, and V.~N. Balasubramanian.
\newblock Grad-cam++: Generalized gradient-based visual explanations for deep
  convolutional networks.
\newblock {\em CoRR}, abs/1710.11063, 2017.

\bibitem{Das2017KeepingTB}
N.~Das, M.~Shanbhogue, S.-T. Chen, F.~Hohman, L.~Chen, M.~E. Kounavis, and
  D.~H. Chau.
\newblock Keeping the bad guys out: Protecting and vaccinating deep learning
  with {JPEG} compression.
\newblock {\em CoRR}, abs/1705.02900, 2017.

\bibitem{Deng2009ImageNetAL}
J.~Deng, W.~Dong, R.~Socher, L.-J. Li, K.~Li, and F.~fei Li.
\newblock Imagenet: A large-scale hierarchical image database.
\newblock {\em 2009 IEEE Conference on Computer Vision and Pattern
  Recognition}, pages 248--255, 2009.

\bibitem{DirtyPODiamond2017}
S.~Diamond, V.~Sitzmann, S.~P. Boyd, G.~Wetzstein, and F.~Heide.
\newblock Dirty pixels: Optimizing image classification architectures for raw
  sensor data.
\newblock {\em CoRR}, abs/1701.06487, 2017.

\bibitem{SUREDonoho1992AdaptingTU}
D.~L. Donoho and I.~Johnstone.
\newblock Adapting to unknown smoothness via wavelet shrinkage.
\newblock 1992.

\bibitem{VISUDonoho1994IdealSA}
D.~L. Donoho and I.~Johnstone.
\newblock Ideal spatial adaptation by wavelet shrinkage.
\newblock 1994.

\bibitem{Dziugaite2016ASO}
G.~K. Dziugaite, Z.~Ghahramani, and D.~M. Roy.
\newblock A study of the effect of {JPG} compression on adversarial images.
\newblock {\em CoRR}, abs/1608.00853, 2016.

\bibitem{DropoutFeinman2017}
R.~Feinman, R.~R. Curtin, S.~Shintre, and A.~B. Gardner.
\newblock Detecting adversarial samples from artifacts.
\newblock {\em CoRR}, abs/1703.00410, 2017.

\bibitem{field1987relations}
D.~J. Field.
\newblock Relations between the statistics of natural images and the response
  properties of cortical cells.
\newblock {\em Josa a}, 4(12):2379--2394, 1987.

\bibitem{Getreuer2012RudinOsherFatemiTV}
P.~Getreuer.
\newblock Rudin-osher-fatemi total variation denoising using split bregman.
\newblock {\em IPOL Journal}, 2:74--95, 2012.

\bibitem{Goodfellow2014ExplainingAH}
I.~J. Goodfellow, J.~Shlens, and C.~Szegedy.
\newblock Explaining and harnessing adversarial examples.
\newblock {\em CoRR}, abs/1412.6572, 2014.

\bibitem{CounteringAIGuo17}
C.~Guo, M.~Rana, M.~Ciss{\'e}, and L.~van~der Maaten.
\newblock Countering adversarial images using input transformations.
\newblock 2017.

\bibitem{He2016DeepRL}
K.~He, X.~Zhang, S.~Ren, and J.~Sun.
\newblock Deep residual learning for image recognition.
\newblock {\em 2016 IEEE Conference on Computer Vision and Pattern Recognition
  (CVPR)}, 2016.

\bibitem{SALICONHuang2015}
X.~Huang, C.~Shen, X.~Boix, and Q.~Zhao.
\newblock Salicon: Reducing the semantic gap in saliency prediction by adapting
  deep neural networks.
\newblock {\em 2015 IEEE International Conference on Computer Vision (ICCV)},
  pages 262--270, 2015.

\bibitem{Hubel1959ReceptiveFO}
D.~H. Hubel and T.~N. Wiesel.
\newblock Receptive fields of single neurones in the cat's striate cortex.
\newblock {\em The Journal of physiology}, 148:574--91, 1959.

\bibitem{ThresholdingJansen2012noise}
M.~Jansen.
\newblock {\em Noise reduction by wavelet thresholding}, volume 161.
\newblock Springer Science \& Business Media, 2012.

\bibitem{Kurakin2016AdversarialEI}
A.~Kurakin, I.~J. Goodfellow, and S.~Bengio.
\newblock Adversarial examples in the physical world.
\newblock {\em CoRR}, abs/1607.02533, 2016.

\bibitem{ForesightLin2017DetectingAA}
Y.-C. Lin, M.-Y. Liu, M.~Sun, and J.-B. Huang.
\newblock Detecting adversarial attacks on neural network policies with visual
  foresight.
\newblock {\em CoRR}, abs/1710.00814, 2017.

\bibitem{Liu2016DelvingIT}
Y.~Liu, X.~Chen, C.~Liu, and D.~X. Song.
\newblock Delving into transferable adversarial examples and black-box attacks.
\newblock {\em CoRR}, abs/1611.02770, 2016.

\bibitem{FoveationbasedMALuo2015}
Y.~Luo, X.~Boix, G.~Roig, T.~A. Poggio, and Q.~Zhao.
\newblock Foveation-based mechanisms alleviate adversarial examples.
\newblock {\em CoRR}, abs/1511.06292, 2015.

\bibitem{Madry2017TowardsDL}
A.~Madry, A.~Makelov, L.~Schmidt, D.~Tsipras, and A.~Vladu.
\newblock Towards deep learning models resistant to adversarial attacks.
\newblock {\em CoRR}, abs/1706.06083, 2017.

\bibitem{Marcelja1980MathematicalDO}
S.~Marc̆elja.
\newblock Mathematical description of the responses of simple cortical cells.
\newblock {\em Journal of the Optical Society of America}, 70 11:1297--300,
  1980.

\bibitem{DeepGazeKmmerer2014DeepGI}
y.~v. Matthias~K{\"{u}mmerer and Lucas Theis and Matthias Bethge},
  journal={CoRR}.
\newblock Deep gaze i: Boosting saliency prediction with feature maps trained
  on imagenet.

\bibitem{Meng2017MagNetAT}
D.~Meng and H.~Chen.
\newblock Magnet: A two-pronged defense against adversarial examples.
\newblock In {\em CCS}, 2017.

\bibitem{MoosaviDezfooli2016DeepFoolAS}
S.-M. Moosavi-Dezfooli, A.~Fawzi, and P.~Frossard.
\newblock Deepfool: A simple and accurate method to fool deep neural networks.
\newblock {\em 2016 IEEE Conference on Computer Vision and Pattern Recognition
  (CVPR)}, pages 2574--2582, 2016.

\bibitem{EasilyFNguyen2015DeepNN}
A.~M. Nguyen, J.~Yosinski, and J.~Clune.
\newblock Deep neural networks are easily fooled: High confidence predictions
  for unrecognizable images.
\newblock {\em 2015 IEEE Conference on Computer Vision and Pattern Recognition
  (CVPR)}, pages 427--436, 2015.

\bibitem{papernot2017cleverhans}
I.~G. R. F. F. F. A. M. K. H. Y.-L. J. A. K. R. S. A. G. Y.-C.~L.
  Nicolas~Papernot, Nicholas~Carlini.
\newblock cleverhans v2.0.0: an adversarial machine learning library.
\newblock {\em arXiv preprint arXiv:1610.00768}, 2017.

\bibitem{WeaklyOquab2015IsOL}
M.~Oquab, L.~Bottou, I.~Laptev, and J.~Sivic.
\newblock Is object localization for free? - weakly-supervised learning with
  convolutional neural networks.
\newblock {\em 2015 IEEE Conference on Computer Vision and Pattern Recognition
  (CVPR)}, pages 685--694, 2015.

\bibitem{papernot2016limitations}
N.~Papernot, P.~McDaniel, S.~Jha, M.~Fredrikson, Z.~B. Celik, and A.~Swami.
\newblock The limitations of deep learning in adversarial settings.
\newblock In {\em Security and Privacy (EuroS\&P), 2016 IEEE European Symposium
  on}. IEEE, 2016.

\bibitem{Papernot2016PracticalBA}
N.~Papernot, P.~D. McDaniel, I.~J. Goodfellow, S.~Jha, Z.~B. Celik, and
  A.~Swami.
\newblock Practical black-box attacks against deep learning systems using
  adversarial examples.
\newblock {\em CoRR}, abs/1602.02697, 2016.

\bibitem{Papernot2016DistillationAA}
N.~Papernot, P.~D. McDaniel, X.~Wu, S.~Jha, and A.~Swami.
\newblock Distillation as a defense to adversarial perturbations against deep
  neural networks.
\newblock {\em 2016 IEEE Symposium on Security and Privacy (SP)}, 2016.

\bibitem{Prakash2017SemanticPI}
A.~Prakash, N.~Moran, S.~Garber, A.~DiLillo, and J.~Storer.
\newblock Semantic perceptual image compression using deep convolution
  networks.
\newblock {\em 2017 Data Compression Conference (DCC)}, 2017.

\bibitem{WaveletDenoisingRangarajan2002}
R.~Rangarajan, R.~Venkataramanan, and S.~Shah.
\newblock Image denoising using wavelets.
\newblock 2002.

\bibitem{Rust2005SpatiotemporalEO}
N.~C. Rust, O.~Schwartz, J.~A. Movshon, and E.~P. Simoncelli.
\newblock Spatiotemporal elements of macaque v1 receptive fields.
\newblock {\em Neuron}, 46:945--956, 2005.

\bibitem{Simoncelli1999BayesianDO}
E.~P. Simoncelli.
\newblock Bayesian denoising of visual images in the wavelet domain.
\newblock 1999.

\bibitem{Su2017OnePA}
J.~Su, D.~V. Vargas, and K.~Sakurai.
\newblock One pixel attack for fooling deep neural networks.
\newblock {\em CoRR}, abs/1710.08864, 2017.

\bibitem{Szegedy2013IntriguingPO}
C.~Szegedy, W.~Zaremba, I.~Sutskever, J.~Bruna, D.~Erhan, I.~J. Goodfellow, and
  R.~Fergus.
\newblock Intriguing properties of neural networks.
\newblock {\em CoRR}, abs/1312.6199, 2013.

\bibitem{IntriguingSzegedy2013}
C.~Szegedy, W.~Zaremba, I.~Sutskever, J.~Bruna, D.~Erhan, I.~J. Goodfellow, and
  R.~Fergus.
\newblock Intriguing properties of neural networks.
\newblock {\em CoRR}, abs/1312.6199, 2013.

\bibitem{Tramr2017EnsembleAT}
F.~Tram{\`e}r, A.~Kurakin, N.~Papernot, D.~Boneh, and P.~D. McDaniel.
\newblock Ensemble adversarial training: Attacks and defenses.
\newblock {\em CoRR}, abs/1705.07204, 2017.

\bibitem{Tramr2017TheSO}
F.~Tram{\`e}r, N.~Papernot, I.~J. Goodfellow, D.~Boneh, and P.~D. McDaniel.
\newblock The space of transferable adversarial examples.
\newblock {\em CoRR}, abs/1704.03453, 2017.

\bibitem{MitigatingAnon208}
C.~Xie, J.~Wang, Z.~Zhang, Z.~Ren, and A.~Yuille.
\newblock Mitigating adversarial effects through randomization.
\newblock In {\em International Conference on Learning Representations}, 2018.

\bibitem{FeatureSqueezingXu2017}
W.~Xu, D.~Evans, and Y.~Qi.
\newblock Feature squeezing: Detecting adversarial examples in deep neural
  networks.
\newblock {\em CoRR}, abs/1704.01155, 2017.

\bibitem{Yosinski2015UnderstandingNN}
J.~Yosinski, J.~Clune, A.~M. Nguyen, T.~J. Fuchs, and H.~Lipson.
\newblock Understanding neural networks through deep visualization.
\newblock {\em CoRR}, abs/1506.06579, 2015.

\bibitem{CAMZhou2016LearningDF}
B.~Zhou, A.~Khosla, {\`A}.~Lapedriza, A.~Oliva, and A.~Torralba.
\newblock Learning deep features for discriminative localization.
\newblock {\em 2016 IEEE Conference on Computer Vision and Pattern Recognition
  (CVPR)}, pages 2921--2929, 2016.

\end{thebibliography}
}

\end{document}